\documentclass[10pt,journal,cspaper,compsoc]{IEEEtran}
\ifCLASSOPTIONcompsoc
\else
\fi
\ifCLASSINFOpdf
\else
\fi

\usepackage{stfloats}

\usepackage{algorithm,algorithmic}
\usepackage{amsmath}
\usepackage{upgreek}
\usepackage{multirow}
\usepackage{times}
\usepackage{epsfig}
\usepackage{graphicx}
\usepackage{amssymb}
\usepackage{url}
\usepackage{textcomp}
\usepackage{bm}
\usepackage{rotating}
\usepackage{booktabs}
\usepackage{mdwlist}
\usepackage{fmtcount}
\usepackage{clrscode}
\usepackage[numbers,square,comma,sort&compress]{natbib}

\usepackage{subfigure} 
\usepackage{wrapfig}
\usepackage{color}

\renewcommand{\Re}{{\mathbb R}}

\belowdisplayskip \abovedisplayskip
\renewcommand{\baselinestretch}{0.98}

\newenvironment{packed_item}{
\begin{itemize}
  \setlength{\itemsep}{1pt}
  \setlength{\parskip}{0pt}
  \setlength{\parsep}{0pt}
}{\end{itemize}}

\newlength{\sectionReduceTop}
\newlength{\sectionReduceBot}
\newlength{\subsectionReduceTop}
\newlength{\subsectionReduceBot}
\newlength{\abstractReduceTop}
\newlength{\abstractReduceBot}
\newlength{\captionReduceTop}
\newlength{\captionReduceBot}
\newlength{\subsubsectionReduceTop}
\newlength{\subsubsectionReduceBot}

\newlength{\horSkip}
\newlength{\verSkip}

\newlength{\figureHeight}
\newcommand{\todo}[1]{\textcolor{black}{\textbf{#1}}}
\newcommand{\todoo}[1]{\textcolor{black}{\textbf{#1}}}

\begin{document}
%
\title{Towards Holistic Scene Understanding: \\Feedback Enabled Cascaded Classification Models}
%
%
%
%

\author{Congcong~Li,~\IEEEmembership{Student Member,~IEEE,}
        Adarsh~Kowdle,~\IEEEmembership{Student Member,~IEEE,}
        Ashutosh~Saxena,~\IEEEmembership{Member,~IEEE,}        
        Tsuhan~Chen,~\IEEEmembership{Fellow,~IEEE}
\IEEEcompsocitemizethanks{\IEEEcompsocthanksitem Congcong Li, Adarsh Kowdle and Tsuhan Chen
are with the School of Electrical and Computer Engineering, Cornell University, Ithaca,
NY, 14853. 
E-mail: \{cl758, apk64\}@cornell.edu, tsuhan@ece.cornell.edu
\IEEEcompsocthanksitem Ashutosh Saxena is with the Department of Computer
Science, Cornell University, Ithaca, NY, 14853. 
E-mail: asaxena@cs.cornell.edu}
\thanks{}}

\IEEEcompsoctitleabstractindextext{%
\begin{abstract}
Scene understanding includes many related sub-tasks, such as scene
categorization, depth estimation, object detection, etc. Each of these sub-tasks is
often notoriously hard, and state-of-the-art classifiers already exist for many
of them.  These classifiers operate on the same raw image and provide 
correlated outputs. It is desirable to have an algorithm that can capture such
correlation without requiring any changes to the inner workings of any
classifier.  

We propose Feedback Enabled Cascaded Classification Models (FE-CCM), 
that jointly optimizes all the sub-tasks,
while requiring only a `black-box' interface to the original classifier for each sub-task.  We use a 
two-layer cascade of classifiers, which are repeated instantiations
of the original ones, with the output of the first layer fed into the
second layer as input.  Our training method involves a feedback step that
allows later classifiers to provide earlier classifiers information about which error
modes to focus on. We show that our method significantly improves performance
in \textit{all} the sub-tasks in the domain of scene understanding, 
where we consider depth estimation, scene categorization, event 
categorization, object detection, geometric labeling and saliency detection.
Our method also improves performance in two robotic applications: 
an object-grasping robot and an object-finding robot.

\end{abstract}

}

\maketitle

\IEEEdisplaynotcompsoctitleabstractindextext

%
\IEEEpeerreviewmaketitle

\section{Introduction}

\begin{figure}[t]
\vspace{-5pt}	
\centering
	 {\includegraphics[width=0.8\linewidth]{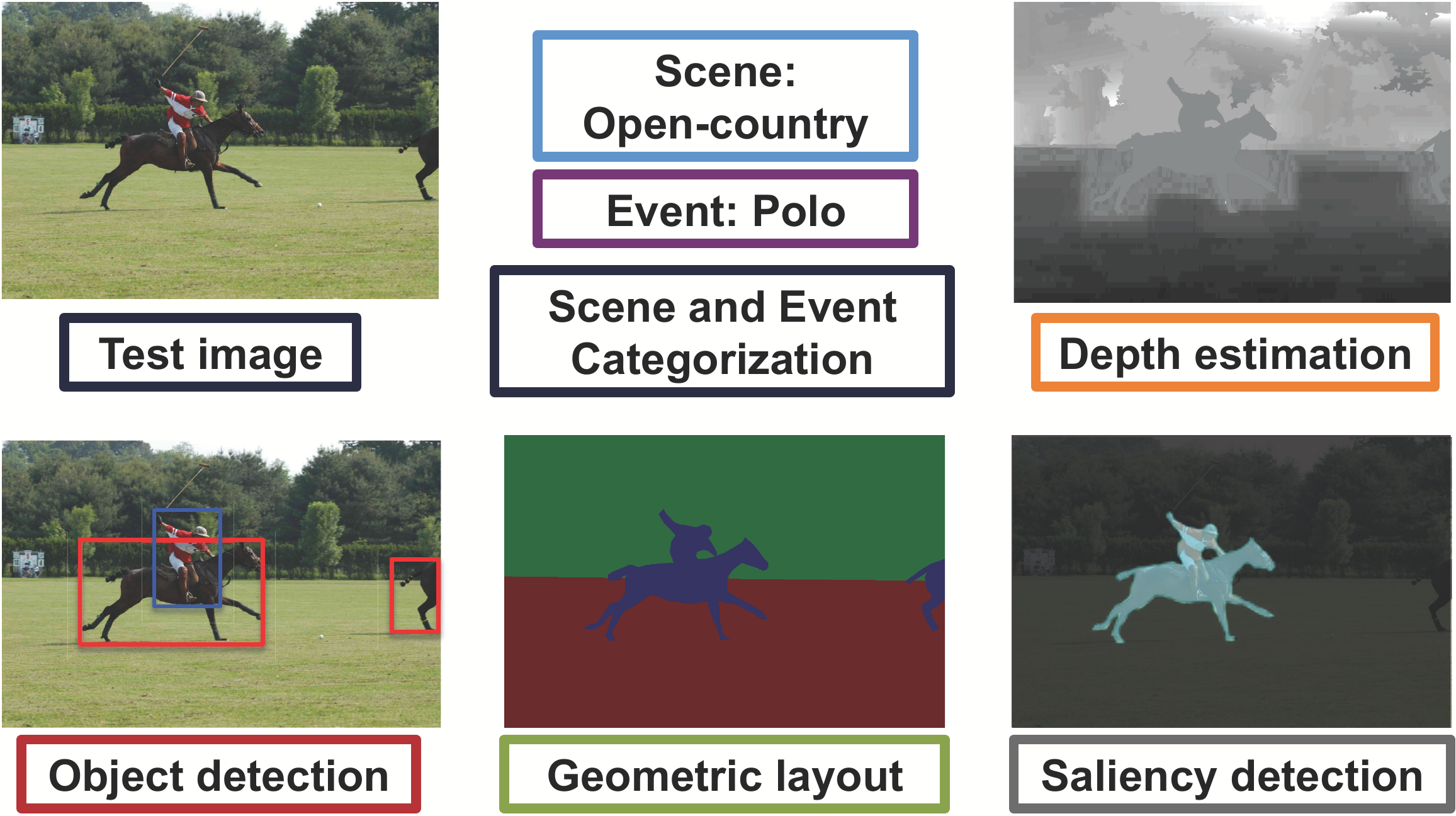}
	 } 
\vspace{-5pt}	 
\caption{\footnotesize{Given a test image, \textit{Holistic Scene Understanding}
corresponds to inferring the labels for all possible scene understanding
dimensions. In our work, we infer labels corresponding to, scene
categorization, event categorization, depth estimation (Black = close, white =
far), object detection, geometric layout (green =
vertical, red = horizontal, blue = vertical) and saliency detection
(cyan = salient) as shown above and achieve this jointly using one unified
model.
Note that different tasks help each other, for example, the
depth estimate of the scene can help the object detector look for the horse;
the object detection can help perform better saliency detection, etc.}}
\vspace{-18pt}	 
\label{fig:scene_und}
\end{figure}

\IEEEPARstart{O}{ne} of the primary goals in computer vision is holistic scene
understanding, which involves many sub-tasks, such as depth estimation, scene
categorization, saliency detection, object detection, event categorization,
etc. (See Figure~\ref{fig:scene_und}.) Each of these tasks explains some aspect of a
particular scene and in order to fully understand a scene, we would need to solve for each of
these sub-tasks.  Several independent efforts have resulted in good classifiers
for each of these sub-tasks. In practice, we see that the sub-tasks are
coupled---for example, if we know that the scene is an indoor scene, it would help us
estimate depth from that single image more accurately.  In another example in
the robotic grasping domain, if we know what kind of object we are trying to grasp, then it is easier
for a robot to figure out how to pick it up.  In this paper, we propose a
unified model that jointly optimizes for all the sub-tasks, allowing them to
share information and guide the classifiers towards a joint optimal.  We show
that this can be seamlessly applied across different applications. 

Recently, several approaches have tried to combine these different classifiers
for related tasks in vision \cite{Kumar_iccv05,Saxena:TPAMI,Hoiem_2008_6034,feifei_CVPR09,Torralba_CVPR06,sutton05,parikh_context_cvpr08,Taskar,Triggs,mrf_stereo}; 
however, most of them tend to be ad-hoc (i.e., a hard-coded rule is used) and
often an intimate knowledge of the inner workings of the individual classifiers
is required.  Even beyond vision, in many other domains, state-of-the-art
classifiers already exist for many sub-tasks.  However, these carefully
engineered models are often tricky to modify, or even to simply 
re-implement from the available descriptions. Heitz
et.~al.~\cite{ashutosh_nips} recently developed a framework for scene
understanding called Cascaded Classification Models ({\bf CCM}) treating each
classifier as a `black-box'. Each classifier is repeatedly instantiated with
the next layer using the outputs of the previous classifiers as inputs.  While
this work proposed a method of combining the classifiers in a way that
increased the performance in all of the four tasks they considered, it
had a drawback that it optimized for each task independently and there was no
way of feeding back information from later classifiers to earlier classifiers
during training. 
This feedback can potentially help the CCM achieve a more optimal
solution.

In our work, we propose Feedback Enabled Cascaded Classification Models 
({\bf FE-CCM}), which provides feedback from the later classifiers 
to the earlier ones, during the training phase. This feedback, provides earlier stages 
information about what error modes should be focused on, or what can be 
ignored without hurting the performance of the later classifiers. 
For example, misclassifying a street scene as highway may not hurt as much as
misclassifying a street scene as open country. Therefore we prefer the first
layer classifier to focus on fixing the latter error instead of optimizing the
training accuracy.  In another example, allowing the depth estimation to focus
on some specific regions can help perform better scene categorization.
For instance, the open country scene is characterized by its upper part as a wide
sky area. Therefore, estimating the depth well in that region by sacrificing
some regions in the bottom may help to correctly classify an image.
In detail, we do so by jointly optimizing all the tasks;
the outputs of the first layers are treated as latent variables
and training is done using an iterative algorithm.   
Another benefit of our method is that each of the classifiers 
can be trained using their own independent training datasets, i.e., our model does
not require a datapoint to have labels for all the sub-tasks, and hence
it scales well with \textit{heterogeneous} datasets.

In our approach, we treat each classifier as a `black-box', with no
restrictions on its operation other than requiring the ability to train
on data and have an input/output interface. (Often each of these individual 
classifier could be quite complex, e.g., producing labelings
over pixels in an entire image.)  Therefore, our method is applicable to many other
tasks that have different but correlated outputs.  

In extensive experiments, we show that our method achieves significant
improvements in the performance of \textit{all} the six sub-tasks we consider:
depth estimation, object detection, scene categorization, event categorization,
geometric labeling and saliency detection. We also successfully apply the same
model to two robotics applications: robotic grasping, and robotic object
detection. 

The rest of the paper is organized as follows. We first define holistic scene understanding
and discuss the related works in Section~\ref{related}. 
We describe our FE-CCM method in Section~\ref{model} followed by the discussion about
handling heterogeneous datasets in Section~\ref{sec:heterogeneous}. We provide the
implementation details of the classifiers 
in Section~\ref{implementation}. We present the experiments and results in
Section~\ref{S.Results} and some robotic applications in Section~\ref{Robotics}. 
We finally conclude in Section~\ref{conclusions}.

\vspace*{\sectionReduceTop}
\section{\label{related}Overview of Scene Understanding}
\vspace*{\sectionReduceBot}

\subsection{Holistic Scene Understanding}
\vspace*{\subsectionReduceBot}

When we look at an image of a scene, such as in Figure~\ref{fig:scene_und}, we 
are often interested in answering several different questions:
What objects are there in the image?  How far are things?  What is going
on in the scene? What type of scene it is?  And so on. 
These are only a few examples of questions in the area of scene understanding;
and there may even be more. 

In the past, the focus has been to address each task in isolation, where
the goal of each task is to produce a label $Y_i \in {\bf S_i}$ for the $i^{th}$
sub-task. If we are considering depth estimation (see Figure \ref{fig:scene_und}), then the
label would be $Y_1 \in {\bf S_1} = \Re_{+}^{100\times 100}$ for continuous values of depth
in a $100 \times 100$ output. For scene categorization, we will have
$Y_2 \in {\bf S_2} = \{ 1, \ldots, K\}$ for $K$ scene classes. 
If we have $n$ sub-tasks, then we would have to produce an output as:
$$\mathcal{Y} = \{ Y_1,\ldots,Y_n \} \in {\bf S_1} \times {\bf S_2} \ldots \times {\bf S_n}.$$
The interesting part here is that often we want to solve different
combinations of the sub-tasks depending on the situation. The goal of this work is 
to design an algorithm that does not depend on the particular sub-tasks
in question.

\vspace*{\subsectionReduceTop}
\subsection{Related Work}
\vspace*{\subsectionReduceBot}

\noindent
\textbf{Cascaded classifiers.}
Using information from related tasks to improve the performance
of the task in question has been studied in various fields of
machine learning. The idea of cascading layers of classifiers to aid
a task was first introduced with neural networks as multi-level
perceptrons where, the output of the first layer of perceptrons is passed on as
input to the next layer \cite{hansen_pami90, shang_icassp93,NeuralDeepNLP}. However,
it is often hard to train neural networks and gain an insight into its operation,
making it hard to work for complicated tasks.  

The idea of improving classification performance by combining outputs of
many classifiers is used in methods such as Boosting \cite{freund_boosting}, 
where many weak learners are combined to obtain a more accurate classifier;
this has been applied to tasks such as face detection
\cite{brubaker_boost_ijcv08, viola_jones_ijcv04}. To incorporate contextual information, Fink and Perona \cite{fink_nips04} exploited local dependencies between objects in a boosting framework, but did not allow for multiple rounds of communication between objects. Torralba et al.~\cite{torralba_nips05} introduced Boosted Random Fields to model object dependency, which used boosting to learn the graph structure and local evidence of a conditional random field. Tu \cite{tu_auto_context_cvpr08} proposed a more general framework which used pixel-level label maps to learn a contextual model through a cascaded classifier approach. All these works mainly consider the interactions between labels of the same type. However, in our CCM framework \cite{feccm_nips,feccm_eccv}, the focus is on capturing contextual interactions between labels of different types. Furthermore, compared to the feed-forward only cascade method in \cite{tu_auto_context_cvpr08}, our model with feedback not only iteratively refines the contextual interactions, but also refines the individual classifiers to provide helpful context.

\noindent
\textbf{Sensor fusion.}
There has been a huge body of work in the area of sensor fusion where
classifiers output the same labels but work with different modalities, each one
giving additional information and thus improving the performance, e.g., 
in biometrics, data from voice recognition and face recognition is combined
\cite{kittler_pami98}. However, in our scenario, we consider multiple 
tasks where each classifier is tackling a different problem 
(i.e., predicting different labels), with the same input being
provided to all the classifiers.

\noindent
\textbf{Structured Models for combining tasks.}
While the methods discussed above combine classifiers to predict the \textit{same} labels, 
there is a group of works 
that designs models for predicting heterogenous labels.
Kumar and Hebert \cite{Kumar_iccv05} developed
a large MRF-based probabilistic model to link multi-class segmentation and
object detection. Li et al. \cite{Li_nips2011} modeled mulitple interactions within tasks and across tasks
by defining a MRF over parameters.
Similar efforts have been made in the field of natural language
processing. Sutton and McCallum \cite{sutton05} combined a parsing model with a semantic role
labeling model into a unified probabilistic framework that solved both
simultaneously. Ando and Zhang \cite{Ando_jmlr05} proposed a general framework for learning predictive functional structures from 
multiple tasks. All these models require knowledge of the inner workings of the individual classifiers, which makes it hard to fit existing state-of-the-art classifiers of certain tasks into the models.

Structured learning algorithms (e.g., \cite{StructuredLearning,
TaskarNIPS2003,structured3Dscene}) can also be a viable option for the setting of combining
multiple tasks. There has been recent development in structured learning on handling latent variables (e.g. hidden conditional random field \cite{Trevor_nips04}, latent structured SVM \cite{Thorsten_icml09}), which can be potentially applied to multi-task settings with disjoint datasets. With 
considerable understanding into each of the tasks, the loss function in structured learning provides a nice way to leverage different tasks. However, in this work, we focus on developing a more generic algorithm that can 
 be easily applied even without intimate knowledge of the tasks. 

There have been many works which show that with a well-designed model,
one can improve the performance of a particular task by using cues 
from other tasks (e.g., \cite{parikh_context_cvpr08,Taskar,Triggs}). 
Saxena et~al. manually designed the terms in an MRF to combine
depth estimation with object detection~\cite{Saxena:TPAMI}
and stereo cues~\cite{mrf_stereo}.
Sudderth et al.~\cite{Torralba_CVPR06} used object recognition to help 3D structure estimation. 

\noindent
\textbf{Context.}
There is a large body of work that leverages contextual information to help
specific tasks. Various sources of context have been explored, ranging from the
global scene layout, interactions between objects and regions to local
features. To incorporate scene-level information, Torralba et al.~\cite{torralba_ijcv03, torralba_pami2002} used the statistics of low-level
features across the entire scene to prime object detection or help depth
estimation. Hoiem et al.~\cite{hoiem_IJCV} used 3D scene information to provide
priors on potential object locations. Park et al.~\cite{ref:RamananECCV2010}
used the ground plane estimation as contextual information for pedestrian
detection. Many works also model context to capture the local interactions
between neighboring regions
\cite{heitz_eccv07,ref:LimICCV2009,ref:KumarICCV2005}, objects \cite{pedro,
ref:RabinovichICCV2007, ref:ParikhCVPR2008, ref:DesaiICCV2009,
YaoFei-FeiContext}, or both \cite{ref:GalleguillosCVPR2010,ref:HoiemCVPR2009,
ref:BlaschkoBMVC2009}. These methods improve the performance of some specific
tasks by combining information from different aspects. However, most of these
methods can not be applied to cases when we only have ``black-box'' classifiers
for the individual tasks.

\noindent
\textbf{Holistic Scene Understanding.}
Hoiem et.~al.~\cite{Hoiem_2008_6034} proposed an innovative but ad-hoc system 
that combined boundary detection and surface labeling
by sharing some low-level information between the classifiers.
Li et.~al.~\cite{feifei_CVPR09,feifei2} combined image classification, annotation and
segmentation with a hierarchical graphical model. 
However, these methods required considerable attention to each classifier, and
considerable insight into the inner workings of each task and also the connections between them.  
This limits the generality of the approaches
in introducing new tasks easily or being applied to other domains.

\noindent
\textbf{Deep Learning.}
There is also a large body of work in the areas of deep learning,
and we refer the reader to
Bengio and LeCun \cite{BengioLeCun2007} for a nice overview of deep learning
architectures and Caruana \cite{MultitaskLearning} for multitask learning with
shared representation. While efficient back-propagation methods like \cite{lecun-98b} 
have been commonly used in learning a multi-layer network, it is not as easy to apply to our case where each node is a complex classifier.  
Most works in deep learning (e.g.,
\cite{Goodfellow,DBN,Fergus}) are different from our work in that, those works focus on
one particular task (same labels) by building different classifier architectures, 
as compared to our setting of \textit{different} tasks with different
labels.   
Hinton et al.~\cite{DBN} used unsupervised learning to obtain an initial
configuration of the parameters.  This provides a good initialization and hence
their multi-layered architecture does not suffer from local minimas 
during optimization.  At a high-level, we can also look at
our work as a  multi-layered architecture (where each node typically produces
complex outputs, e.g., labels over the pixels in the image); and initialization in our
case comes from existing state-of-the-art individual classifiers.  
Given this initialization, our training procedure finds
parameters that (consistently) improve performance across all the sub-tasks.

\begin{figure}[t]
\vspace{-15pt}
\centering
	 {\includegraphics[width=1.0\linewidth]{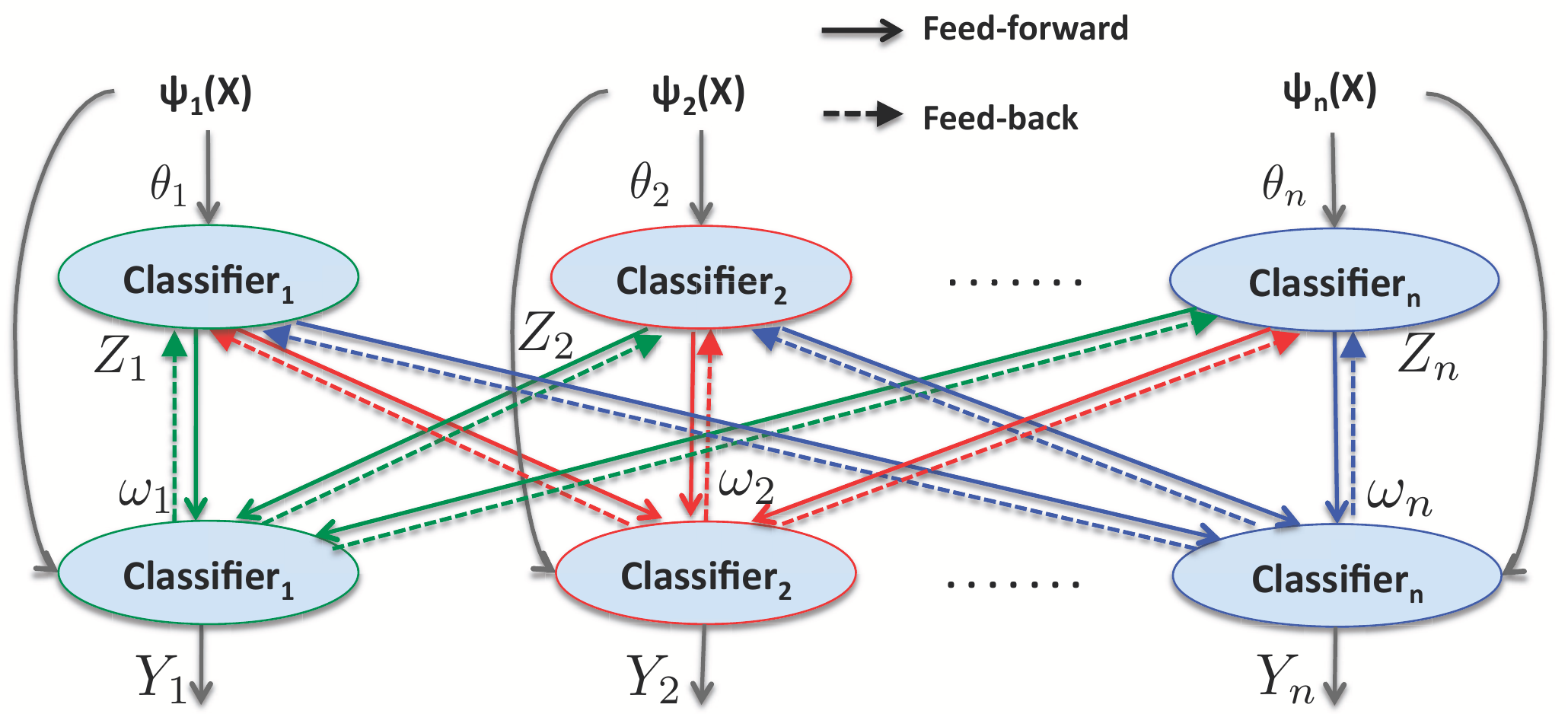}
	 } \vspace{-22pt}
\caption{\footnotesize{The proposed Feed-back enabled cascaded classification model (FE-CCM) 
for combining related classifiers.
($\forall i \in \{1, 2, \dots, n\}$, $\Psi_i(X)$ = Features corresponding to $Classifier_i$ extracted from
image $X$, $Z_i$ = Output of the $Classifier_i$ in the first stage parameterized by
$\theta_i$, $Y_i$ = Output of the $Classifier_i$ in the second stage
parameterized by $\omega_i$). In the proposed FE-CCM model, there is feed-back from the
latter stages to help achieve a model which optimizes all the tasks
considered, jointly. Here $Classifier_i$'s on the two layers can have different forms though they are for the same classification task.  (Note that different colors of lines are used only to make
the figure more readable.)}}
\vskip -.25in
\label{fig:graph}
\end{figure}

\vspace*{\sectionReduceTop}
\section{\label{model}Feedback Enabled Cascaded Classification Models}
\vspace*{\sectionReduceBot}
In the field of scene understanding, a lot of independent research into each of
the vision sub-tasks has led to excellent classifiers. These independent
classifiers are typically trained on different or \emph{heterogenous} datasets
due to the lack of ground-truth labels for all the sub-tasks. 
In addition, each of these classifiers come with their own learning
and inference methods. Our goal is to consider each of them
as a `black-box', which makes it easy to combine them.
We describe what we mean by `black-box classifiers' below.

\smallskip
\noindent\textbf{Black-box Classifier.} 
A black-box classifier, as the name suggests, is a classifier for which 
operations (such as learning and inference algorithms) are available for use,
but their inner workings are not known. We assume that,
given the training dataset 
${\bf X}$, features extracted $\mathrm{\bf \Psi(X)}$ and the target outputs of the
$i^{th}$ task $\mathrm{\bf Y_{i}}$, the black-box classifier has some internal
learning function $f_{\textrm{learn}}^{i}$ with parameters $\theta_{i}$ that optimizes the
mapping from the inputs to the outputs for the training data. \footnote{Unless specified, the 
regular symbols (e.g. $X$, $Y_i$, etc.) are used for a particular data-point, and the bold-face symbols (e.g. ${\bf X}$, ${\bf Y_i}$, etc.) are used for a dataset.}
Once the parameters have been learnt, given a new data point, $X$ with features $\Psi(X) \in \mathbb{R}^{K}$,
where $K$ can be changed
as desired,\footnote{If the input dimension of the black-box classifier can not
be changed, then we will use that black-box in the first layer only.} the
black-box classifier returns the output $\hat{Y_{i}}$ according to its internal
inference function $f_{\textrm{infer}}^{i}$. This is illustrated through the equations
below. For the $i^{th}$ task,
\begin{equation}
\small{
\text{\emph{Learning}}: \theta_{i}^* = \operatorname*{optimize}_{\theta_{i}} f_{\textrm{learn}}^{i}(\mathrm{\bf \Psi(X)},\mathrm{\bf Y_{i}}; \theta_{i})\label{eq.blackboxlearn}
}
\end{equation}
\begin{equation}
\small{
\text{\emph{Inference}}: \hat{Y_{i}} = \operatorname*{optimize}_{Y_{i}} f_{\textrm{infer}}^{i}(\Psi(X), Y_{i}; \theta_{i}^*)\label{eq.blackboxinfer}.
}
\end{equation}
This approach of treating each classifier as a black-box allows us to use
different existing classifiers which have been known to perform well at
specific sub-tasks. Furthermore, without changing their inner workings,  
it allows us to compose them into one model which exploits the information from
each sub-task to aid holistic scene understanding.

\vspace*{\subsectionReduceTop}
\subsection{Our Model}
\label{sec:model}
\vspace*{\subsectionReduceBot}

Our model is built in the form of a two-layer cascade, as shown in
Figure~\ref{fig:graph}. The first layer consists of an instantiation of each of
the black-box classifiers with the image features as input. The second layer is
a repeated instantiation of each of the classifiers with the first layer
classifier outputs as well as the image features as inputs. 
Note that the repeated classifier on the second layer is not necessary to have the same mathematical form with the one on the first layer. Instead, we consider it as a repeated instantiation only because they are used for the same classification task.

\smallskip
\noindent
\textbf{Notation:}
We consider $n$ related sub-tasks ${\rm Classifier}_i$, 
$i \in \{1, 2, \dots, n\}$ (Figure \ref{fig:graph}).
We describe the notations used in this paper as follows:

\begin{table}[h]
\vskip -10pt
\centering
\begin{tabular}{p{1.6cm}p{6cm}}
\hline
\centering{$\Psi_i(X)$} & Features corresponding to ${\rm Classifier}_i$ extracted from image $X$. \\

\centering{$Z_i$, $\mathcal{Z}$} & $Z_{i}$ indicates output from the first layer ${\rm Classifier}_i$. Many classifiers output continuous scores instead of labels. In cases where this is not the case, it is trivial to convert a binary classiÞerÕs output to a log-odds scores. For a $K$-class ($K>2$) classifier, we consider the output to be a $K$-dimensional vector. 
$\mathcal{Z}$ indicates the set $\{Z_1, Z_2, \dots, Z_n\}$.\\

\centering{$\theta_i$, $\Theta$} & $\theta_i$ indicates parameters corresponding to first layer ${\rm Classifier}_i$. $\Theta$ indicates the set $\{\theta_1,\dots,\theta_n\}$. \\ 

\centering{$Y_j$, $\mathcal{Y}$} & $Y_j$ indicates output for the $j^{th}$ task in the second layer, using the original features $\Psi_j(X)$ as well as all the outputs from the first layer as input. $\mathcal{Y}$ indicates the set $\{Y_1, Y_2, \dots, Y_n\}$. \\

\centering{$\omega_j$, $\Omega$} &  $\omega_j$ indicates parameters for the second layer ${\rm Classifier}_j$. $\Omega$ indicates the set $\{\omega_1,\dots,\omega_n\}$.\\

\centering{$\Gamma_j$, $\Gamma$} & Dataset for the $j^{th}$ task, which consists of labeled pairs $\{X,Y_j\}$ in the training set. $\Gamma$ represents all
the labeled data.  \\

\centering{${f}_{\textrm{infer}}^{i}$, ${f}_{\textrm{learn}}^{i}$} & the internal inference function and learning function for the $i^{th}$ classifier on the first layer.\\

\centering{${f'}_{\textrm{infer}}^{i}$, ${f'}_{\textrm{learn}}^{i}$} & the internal inference function and learning function for the $i^{th}$ classifier on the second layer.\\

\hline
\end{tabular}\\
\vskip -10pt
\end{table}

With the notations in place we will now first describe the inference and learning algorithms for the proposed model in the following sections, followed by probabilistic interpretation of our method.

\vspace*{\subsectionReduceTop}
\subsection{Inference Algorithm}
\label{sec:inference}
\vspace*{\subsectionReduceBot}

During inference, the inputs $\Psi_i(X)$ are given and our goal is to infer the 
final outputs $Y_i$.  Using the learned parameters $\theta_i$ for the first level of 
classifiers and $\omega_i$ for the second level of classifiers,
we first infer the first-layer outputs $Z_i$ and then infer the second-layer outputs $Y_i$. 
More formally, we perform the following. 
\begin{equation}
\small{
\hat{Z_{i}} = \operatorname*{optimize}_{Z_{i}} f_{\textrm{infer}}^{i}(\Psi_{i}(X), Z_{i}; \hat{\theta}_{i})\label{eq.inferlayer1}
}
\end{equation}
\begin{equation}
\small{
\hat{Y_{i}} = \operatorname*{optimize}_{Y_{i}} {f'}_{\textrm{infer}}^{i}([\Psi_{i}(X) \; \mathcal{\hat{Z}}], Y_{i}; \hat{\omega}_{i})\label{eq.inferlayer2}
}
\end{equation}
The inference algorithm is given in Algorithm \ref{inference_table}.
This method allows us to use the internal inference function (Equation~\ref{eq.blackboxinfer}) of the black-box classifiers without knowing its inner workings. 
Note that the complexity here is no more than constant times the complexity of inference in the original classiÞers. 
\vspace{-8pt}
\begin{algorithm}
\caption{Inference}
\label{inference_table}
\begin{footnotesize}
1. \textbf{Inference for first layer:}
\begin{list}{\labelitemi}{\leftmargin=1em}
\item [] for $i=1:n$
\item [] \quad Infer the outputs of the $i^{th}$ classifier using Equation~\ref{eq.inferlayer1};
\item [] end
\end{list}

2. \textbf{Inference for second layer:}
\begin{list}{\labelitemi}{\leftmargin=1em}
\item [] for $i=1:n$
\item [] \quad Infer the outputs of the $i^{th}$ classifier using Equation~\ref{eq.inferlayer2};
\item [] end
\end{list}
\end{footnotesize} 
\end{algorithm}
\vspace{-10pt}

\vspace*{\subsectionReduceTop}
\subsection{Learning Algorithm}
\label{sec:learning}
\vspace*{\subsectionReduceBot}

During the training stage, the inputs $\Psi_i(X)$ as well as the target outputs, $Y_1,Y_2,\dots, Y_n$ of the second level of classifiers, are all observed (because the ground-truth labels 
are available). In our algorithm, we consider $\mathcal{Z}$ (outputs of layer 1 and inputs
to layer 2) as hidden variables. In previous work, Heitz et al.~\cite{ashutosh_nips} assume
that each layer is independent and that each layer produces the
best output independently (without consideration for other layers), and therefore
use the ground-truth labels for $\mathcal{Z}$ even for training the classifiers
in the first layer.

On the other hand, we want to optimize for  the final outputs as much as possible. 
Thus the first layer classifiers need not perform their best (w.r.t. 
groundtruth), but rather focus on error modes that would
result in the second layer's output ($Y_1,Y_2,\dots, Y_n$) being more correct.
Therefore, we learn the model through an iterative Expectation-Maximization formulation, given
the independencies between classifiers represented by the model in
Figure~\ref{fig:graph}. In one step (Feed-forward step) 
we assume the variables $Z_i$'s are known and learn the parameters and 
in the other step (Feed-back step) we fix the parameters estimated previously and 
estimate the variables $Z_i$'s. Since the $Z_i$'s are not fixed to the ground truth, as the iterations progress, the first level of classifiers start focusing on the error modes which would give the best improvement in performance at the end of the second level of classifiers. 
The learning algorithm is summarized in Algorithm~\ref{learning_table}.  
\vspace{-10pt}
\begin{algorithm}
\caption{Learning}
\label{learning_table}
\begin{footnotesize} 

1. Initialize latent variables $\mathcal{Z}$ with the ground-truth $\mathcal{Y}$.\\
 
2. Do until convergence or maximum iteration: \{

\quad \textbf{Feed-foward step:} Fix latent variables $\mathcal{Z}$, estimate the parameters $\Theta$ and $\Omega$ using Equation~\ref{eq.forwardlayer1.blackbox} and Equation~\ref{eq.forwardlayer2.blackbox}.

\quad \textbf{Feed-back step:} Fix the parameters $\Theta$ and $\Omega$, compute latent variables $\mathcal{Z}$ using Equation~\ref{eq.feedback}.\\
\}
\end{footnotesize} 
\end{algorithm}
\vspace*{-30pt}

\noindent\textbf{Initialization:}
We initialize the model by setting the latent variables $Z_i$'s
to the groundtruth, i.e. $Z_{i} = Z_{i}^{gt}$.  Training with this initialization, our cascade is equivalent to
CCM in \cite{ashutosh_nips}, where the classifiers (and the parameters) 
in the first layer are similar to the original state-of-the-art
classifier and the classifiers in the second layer use the outputs of the first layer 
in addition to the original features as input.

\noindent
{\bf Feed-forward Step:}
In this step, we estimate the parameters $\Theta$ and $\Omega$. We assume that the 
latent variables $Z_i$'s are known (and $Y_i$'s are known because they are the
ground-truth during learning, i.e. $Y_{i}=Y_{i}^{gt}$).
We then learn the parameters of each classifier independently.
Learning $\theta_{i}$ is precisely the
learning problem of the `black-box classifier', and learning $\omega_{i}$ is also an
instantiation of the original learning problem, but with the original input features appended with the outputs of the first level classifiers. Therefore, we can use the learning
method provided by the individual black-box classifier (Equation~\ref{eq.blackboxlearn}).
\vskip -10pt
\small{
\begin{align}
&\hat{\theta}_{i} = \operatorname*{optimize}_{\theta_{i}} f_{\textrm{learn}}^{i}({\bf\Psi_{i}(X)},{\bf Z}_{i}; \theta_{i})\label{eq.forwardlayer1.blackbox}\\
&\hat{\omega}_{i} = \operatorname*{optimize}_{\omega_{i}} {f'}_{\textrm{learn}}^{i}([{\bf\Psi_{i}(X)} \; \bm{\mathcal{Z}}], {\bf Y}_{i}; \omega_{i})\label{eq.forwardlayer2.blackbox}
\end{align}
}
\normalsize{}
We now have the parameters for all the classifiers.

\noindent
{\bf Feed-back Step:}
In the second step, we will estimate the values of the variables
$Z_i$'s assuming that the parameters are fixed (and $Y_i$'s are
given because the ground-truth is available, i.e. $Y_{i}=Y_{i}^{gt}$).  This
feed-back step is the crux that provides information to the first-layer
classifiers what error modes should be focused on and what can be ignored
without hurting the final performance. Given $\theta_{i}$'s and $\omega_{i}$'s are fixed,
we want the $Z_i$'s to be good predictions from the first-layer classifiers and also help 
to increase the correction predictions of $Y_{i}$'s as much as possible. 
We optimize the following function for the feed-back step:  
\begin{equation}
\small{
\label{eq.feedback}
\operatorname*{optimize}_{\mathcal{Z}}
  \sum_{i=1}^n \left( J_{1}^{i}(\Psi_{i}(X), Z_{i}; \hat{\theta}_i) + J_2^{i}(\Psi_i(X),\mathcal{Z},Y_{i}; \hat{\omega}_i)\right)
}
\end{equation}
\vskip -5pt
where $J_{1}^{i}$'s and $J_{2}^{i}$'s are functions respectively related to the first-layer classifiers and the second-layer classifiers. 
one option is to have $J_{1}^{i}(\Psi_{i}(X), Z_{i}; \hat{\theta}_i) = f_{\textrm{infer}}^{i}(\Psi_{i}(X), Z_{i}; \hat{\theta}_{i})$ and $J_2^{i}(\Psi_{i}(X), \mathcal{Z}, Y_{i}; \hat{\omega}_{i}) = {f'}_{\textrm{infer}}^{i}([\Psi_{i}(X), \mathcal{Z}], Y_{i}; \hat{\omega}_{i})$ if the intrinsic inference functions for the classifiers are known. More discussions will be given in Section \ref{sec:probabilistic} if the intrinsic functions are unknown. The updated $Z_{i}$'s will be used to re-learn the classifier parameters in the feed-forward step of next iteration. Note that the updated $Z_{i}$'s have continuous values. If the internal learning function of a classifier accepts only labels, we threshold the values of $Z_{i}$'s to get labels.

\vspace*{\subsectionReduceTop}
\subsection{Probabilistic Interpretation}
\label{sec:probabilistic}
\vspace*{\subsectionReduceBot}

Our algorithm can be explained with a probabilistic interpretation where the goal is to maximize the log-likelihood of the outputs of all tasks given the observed inputs,
i.e., $\log P(\mathcal{Y} | X)$, where $X$ is an image belonging to training set $\Gamma$. Therefore, the goal of the proposed model shown in Figure~\ref{fig:graph} is to maximize
\begin{equation}
\small{
\log \prod_{X \in \Gamma} P(\mathcal{Y} | X; \Theta, \Omega)
\label{equ:opt}
}
\end{equation}
To introduce the hidden valuables $Z_{i}$'s, we expand Equation~\ref{equ:opt} as follows, using
the independencies represented by the directed model in
Figure~\ref{fig:graph}. 
\vskip -8pt
\small{
\begin{align}
= &\sum_{X \in \Gamma} \log \!\!\sum_{\mathcal{Z}} P(Y_1,\dots,Y_n, \mathcal{Z} | X; \Theta, \Omega)  \\ 
= &\sum_{X \in \Gamma} \log \!\!\sum_{\mathcal{Z}} \prod_{i=1}^n P(Y_i|\Psi_{i}(X), \mathcal{Z}; \omega_{i}) P(Z_i|\Psi_{i}(X); \theta_{i})  
\label{loglikelihood}
\end{align}
}
\normalsize{}
However, the summation inside the $\log$ makes it difficult to learn the
parameters. Motivated by the Expectation Maximization algorithm~\cite{Dempster_EM}, 
we iterate between the two steps as described in the following. Again 
we initialize the classifiers by learning the classifiers with ground-truth
as discussed Section \ref{sec:learning}.  

\noindent
{\bf Feed-forward Step:}
In this step, we estimate the parameters by assuming that the latent
variables $Z_i$'s are known (and $Y_i$'s are known anyway because they are the
ground-truth). This results in
\begin{equation}
\small{
\begin{aligned}
\operatorname*{maximize}_{\theta_1, \dots, \theta_n,\omega_1, \dots, \omega_n} &
\sum_{X \in \Gamma} \log \prod_{i=1}^n P(Y_i|\Psi_{i}(X), \mathcal{Z}; \omega_{i})P(Z_i|\Psi_{i}(X); \theta_{i})  
\label{eq.feedforward}
\end{aligned}
}
\end{equation}

Now in this feed-forward step, the terms for maximizing the different
parameters turn out to be independent. So, for the $i^{th}$ classifier we have:

\vskip -10pt
\small{
\begin{align}
\operatorname*{maximize}_{\theta_{i}} \sum_{X \in \Gamma} & \log P(Z_{i}|\Psi_i(X);\theta_{i})\label{eq.forwardlayer1}\\
\operatorname*{maximize}_{\omega_{i}} \sum_{X \in \Gamma} & \log P(Y_i|\Psi_i(X),\mathcal{Z};\omega_{i})
\label{eq.forwardlayer2}
\end{align}
}
\normalsize{}
Note that the optimization problem nicely breaks down into the
sub-problems of training the individual classifier for the
respective sub-tasks. We can solve each sub-problem separately given the probabilistic interpretation of the corresponding classifier. When the classifier is taken as `black-box', this can be approximated using the original learning method provided by the individual black-box classifier 
(Equation~\ref{eq.forwardlayer1.blackbox} and Equation~\ref{eq.forwardlayer2.blackbox})

\noindent
{\bf Feed-back Step:}
In this step, we estimate the values of the latent variables
$Z_i$'s assuming that the parameters are fixed.  We perform MAP inference on $Z_i$'s (and not marginalization).  This can be considered as a special variant of the general EM framework (hard EM, \cite{neal1998view}). Using Equation~\ref{loglikelihood}, we get the following optimization problem:
\begin{equation}
\small{
\label{feedback}
\begin{aligned}
&\operatorname*{maximize}_{\mathcal{Z}}  \log P(Y_1,\dots,Y_n,\mathcal{Z} | X;\hat{\theta}_{1}, \dots, \hat{\theta}_{n},\hat{\omega}_{1}, \dots, \hat{\omega}_{n}) \Leftrightarrow \\
&\operatorname*{maximize}_{\mathcal{Z}}
  \sum_{i=1}^n \Big( \log P(Y_{i}|\Psi_i(X),\mathcal{Z}; \hat{\omega}_i) + \log P(Z_{i}|\Psi_i(X); \hat{\theta}_i) \Big) 
\end{aligned}
}
\end{equation}
This maximization problem requires that we have access to the characterization
of the individual black-box classifiers in a probabilistic form. 
If the probabilistic interpretations of the classifiers are known, we can solve the above function accordingly. Note that
Equation~\ref{feedback} is same as Equation~\ref{eq.feedback} with $J_{1}^{i}(\Psi_i(X),Z_{i}; \hat{\theta}_i) =\log P(Z_{i}|\Psi_i(X); \hat{\theta}_i)$ and $J_{2}^{i}(\Psi_i(X),\mathcal{Z},Y_{i}; \hat{\omega}_i) = \log P(Y_{i}|\Psi_i(X),\mathcal{Z}; \hat{\omega}_i)$.

In some cases, the classifier log-likelihoods in Equation~\ref{feedback} actually
turn out to be convex. For example, if the individual classifiers 
are linear or logistic classifiers, the minimization problem is convex and can
be solved using gradient descent (or any such method).

However, if the probabilistic interpretations of the classifiers are unknown, the feedback step requires extra modeling. Some modeling options are provided as follows:

\begin{list}{\labelitemi}{\leftmargin=1em}

\item Case 1: Insight into the vision problem is available.
In this case, one could use the domain knowledge of the task into the problem to properly model $J_{1}^{i}$'s and $J_{2}^{i}$'s. 

\item Case 2: No insight into the vision problem is available and no internal function
of the original classifier is known. In this case, we formulate the $J_{1}^{i}$'s and $J_{2}^{i}$'s as follows. The $J_{1}^{i}$ is defined to be a distance function between the target $Z_{i}$ and the estimated $\hat{Z}_{i}$, which serves as a regularization for the first-layer classifiers. 
\begin{equation}
\small{
\begin{aligned}
J_{1}^{i}(\Psi_{i}(X),Z_{i};\hat{\theta}_{i}) = \left|\left|Z_{i}-\hat{Z}_{i}\right|\right|^{2} \\
\textrm{s.t. } \hat{Z}_{i}= \operatorname*{optimize}_{Z_{i}}f_{\textrm{infer}}^{i}(\Psi_i(X),Z_{i};\hat{\theta}_{i})
\end{aligned}
}
\end{equation}
To formulate $J_{2}^{i}$'s, we make a variational approximation 
on the output of the second-layer classifier for task $i$ (i.e., approximating it as a Gaussian, \cite{gibbs_variational}) to get:
\begin{equation}
\small{
\begin{aligned}
&\operatorname*{minimize}_{\alpha_i} \sum_{X \in \Gamma}  \left | \left |\hat{Y_i} - \alpha_i^T [\Psi_i(X), \hat{\mathcal{Z}}]  \right|\right|_2^{2}   
\label{eq.approximate}\\
\end{aligned}
}
\end{equation}
where $\alpha_{i}$ are parameters of the approximation model. $\hat{Y_{i}}$ is the actual output of the second layer classifier for the task $i$, i.e. $\hat{Y}_i =  \operatorname*{optimize}_{Y_{i}} {f'}_{\textrm{infer}}([\Psi_{i}(X) \; \mathcal{\hat{Z}}], Y_{i}; \hat{\omega}_{i})$. Then we define the $J_{2}^{i}$'s as follows.
\begin{equation}
\small{
\begin{aligned}
J_{2}^{i}(\Psi_i(X),\mathcal{Z},Y_{i}; \hat{\omega}_i) = \left | \left |Y_i - \hat{\alpha}_i^T [\Psi_i(X), \mathcal{Z}]  \right|\right|_2^{2}
\end{aligned}
}
\end{equation}

\textbf{Sparsity:}
Note that the parameter $\alpha_{i}$ is typically extremely
high-dimensional (and increases with the number of tasks) because the second
layer classifiers take as input the original features as well as outputs of all
previous layers. The learning for the approximation model may become
ill-conditioned. 
Therefore, we want our model to select only a few non-zero weights, i.e.,
only a few non-zero entries in $\alpha_{i}$. We do this by introducing the $l_{1}$
sparsity in the parameters~\cite{bach_l1}. So Equation~\ref{eq.approximate} is extended as follows.  
\begin{equation}
\small{
\operatorname*{minimize}_{\alpha_i} \sum_{X \in \Gamma} \left ( \left | \left |\hat{Y_i} - \alpha_i^T [\Psi_i(X), \hat{\mathcal{Z}}]  \right|\right|_2^{2}  + \beta \left |\alpha_i \right| \right )  
}
\end{equation}

\end{list}

\noindent \textbf{Inference:} As introduced in Section~\ref{sec:inference}, 
our inference procedure consists of two steps: first maximize over 
hidden variable $Z$ and then maximize over $Y$. 
\footnote{Another alternative would have been to maximize
$P(Y|X) = \sum_Z P(Y,Z|X)$; however, this would require 
marginalization over the variable $Z$ which is expensive to compute.}
\small{
\begin{align}
	\mathcal{\hat{Z}} = \operatorname*{argmax}_{\mathcal{Z}} \log P(\mathcal{Z}|X,\hat{\Theta})\label{eq:inference_layer1}\\
	\mathcal{\hat{Y}} = \operatorname*{argmax}_{\mathcal{Y}} \log P(\mathcal{Y} |\mathcal{\hat{Z}}, X, \hat{\Omega})\label{eq:inference_layer2}
\end{align}
}
\normalsize{}
Given the structure of our directed graph, the outputs for different
classifiers on the same layer are independent given their inputs and
parameters. Therefore, Equations \ref{eq:inference_layer1} and
\ref{eq:inference_layer2} are equivalent to the following: 
\vskip -10pt
\small{
\begin{align}
	\hat{Z_{i}} = \operatorname*{argmax}_{Z_{i}} \log P(Z_{i}|\Psi_{i}(X);\hat{\theta_{i}})\label{eq:inferenceZ}, i = 1,\dots, n\\
	\hat{Y_{i}} = \operatorname*{argmax}_{Y_{i}} \log P(Y_{i} |\Psi_{i}(X);\mathcal{\hat{Z}}; \hat{\omega_{i}}), i = 1,\dots, n\label{eq:inferenceY}	
\end{align}
}
\normalsize{}
As we see, Equation~\ref{eq:inferenceZ} and Equation~\ref{eq:inferenceY} are instantiations of Equation~\ref{eq.inferlayer1} and Equation~\ref{eq.inferlayer2} in the probabilistic form.

\vspace*{-5pt}
\vspace*{\sectionReduceTop}
\section{Training with Heterogeneous datasets}
\vspace*{\sectionReduceBot}
\label{sec:heterogeneous}

Often real datasets are disjoint for different tasks, i.e, each datapoint does
not have the labels for all the tasks. Our formulation handles this scenario well.
In this section, we show our formulation for this general case, where we use $\Gamma_i$ as the
dataset that has labels only for the $i^{th}$ task. 

In the following we provide the modifications to the feed-forward step and the feed-back step while
dealing with disjoint datasets, i.e., data in dataset $\Gamma_i$ only have labels for the $i^{th}$ task. These modifications also allow us to develop different variants of the model, described
in Section~\ref{sec:variants}.

\noindent \textbf{Feed-forward Step:}
Using the feedback step, we can have $Z_{i}$'s for all the data. Therefore, we use all the datasets in order to re-learn each of the first-layer classifiers. If the internal learning function of the black-box classifier is additive over the data points, then we  have 
\begin{equation}
\small{
\hat{\theta}_{i} = \operatorname*{optimize}_{\theta_{i}} \sum_{j}\sum_{X\in\Gamma_{j}}\pi_{j}f_{\textrm{learn}}^{i}(\Psi_{i}({X}), Z_{i}; \theta_{i}),
\label{eq.forwardlayer1_heter}
}
\end{equation}
where $\pi_{j}$'s are the importance factors given to different datasets, and satisfy $\sum_{j} \pi_{j} = 1$. (See Section~\ref{sec:variants} on how to choose $\pi_j$'s.)

If the internal learning function is not additive over the data points, we provide an alternative solution here. We sample a subset of data ${\bf X}^{j}$ from each dataset $\Gamma^{j}$, i.e. ${\bf X}^{j}\subseteq \Gamma^{j}$ and combine them into a new set ${\bf X} = [{\bf X}^{1}, \dots, {\bf X}^{n}]$. In ${\bf X}$, the ratio of data belonging to ${\bf X}^{j}$ is equal to $\pi_{j}$, i.e. $\frac{|{\bf X}^{j}|}{|{\bf X}|}=\pi_{j}$, where $|\cdot |$ indicates the number of data-points. 
Then we can learn the parameters of the first-layer classifiers as follows.
\begin{equation}
\small{
\hat{\theta}_{i} = \operatorname*{optimize}_{\theta_{i}} f_{\textrm{learn}}^{i}({\bf \Psi_{i}(X)},{\bf Z_{i}}; \theta_{i}),
\label{{eq.forwardlayer1_heter_alter}}
}
\end{equation}
To re-learn the second-layer classifiers, the only change made to Equation~\ref{eq.forwardlayer2.blackbox} is that instead of using all data while optimizing for a particular task, we use only the data-points that have the ground-truth label for the corresponding task.  
\begin{equation}
\small{
\hat{\omega}_{i} = \operatorname*{optimize}_{\omega_{i}} {f'}_{\textrm{learn}}^{i}([{\bf \Psi_{i}(X)} \; \bm{\mathcal{Z}}], {\bf Y}_{i}; \omega_{i})\label{{eq.forwardlayer2_heter}}, \textrm{ s.t. } {\bf X} = \Gamma_{i}
}
\end{equation}

\noindent \textbf{\todo{Feed-back Step}:} In this step, we change Equation~\ref{eq.feedback} as follows. Since a datapoint in the set $\Gamma_{j}$ only has ground-truth label for the $j^{th}$ task ($Y_{j}$), 
we only consider $J_{2}^{j}$ in the second term. However, since this datapoint has outputs for all the first-layer classifiers using the feed-forward step, we consider all the $J_{1}^{i}$'s, $i=1,\cdots,n$. Therefore, in order to obtain the value of $\mathcal{Z}$ corresponding to each data-point $X\in \Gamma_{j}$, we have
\begin{equation}
\small{
\label{eq.feedback_heter}
\operatorname*{optimize}_{\mathcal{Z}}
  \sum_{i=1}^n \big(J_{1}^{i}(\Psi_j(X),Z_{i};\hat{\theta}_i)\big) + J_{2}^{j}(\Psi_j(X),\mathcal{Z},Y_{j};\hat{\omega}_j).
  }
\end{equation}

\vskip -.1in
\vspace*{\subsectionReduceTop}
\subsection{FECCM: Different Instantiations\label{sec:variants}}
\vspace*{\subsectionReduceBot}
The parameters $\pi_j$ allow us to formulate three different instantiations of our model. 
\begin{list}{\labelitemi}{\leftmargin=1em}
\item {\bf Unified FECCM}: In this instantiation, our goal is to achieve improvements in all tasks with one set of parameters $\{\Theta, \Omega\}$. We want to balance the data from different datasets (i.e., with different task labels). Towards this goal, $\pi_j$ is set to be inversely proportional to the amount of data in the dataset of the $j^{th}$ task. Therefore, the unified FECCM balances the amount of data in different datasets, based on Equation~\ref{eq.forwardlayer1_heter}.

\medskip
\item {\bf One-goal FECCM}: In this instantiation, we set $\pi_j = 1$ if
$j = k$, and $\pi_j = 0$ otherwise. This is an extreme setting to
favor the specific task $k$. In this case, the retraining of the first-layer
classifiers will only use the feedback from the ${\rm Classifier}_{k}$ on the second
layer, i.e., only use the dataset with labels for the $k^{th}$ task. Therefore,
FECCM degrades to a model with only one target task (the $k^{th}$ task) on the
second layer and all the other tasks are only instantiated on the first layer.
Although the goal in this setting is to completely benefit the $k^{th}$ task, in practice
it often results in overfitting and does not always achieve the best results
even for the specific task (see Table \ref{table:table_results} in Section
\ref{S.Results}). In this case, we train different models, i.e. different $\theta_{i}$'s and $\omega_{i}$'s, for different target tasks.

\medskip
\item {\bf Target-Specific FECCM}: This instantiation is to
optimize the performance of a specific task. 
As compared to one-goal FECCM
where we manually remove the other tasks on the second layer, in this instantiation
we keep all the
tasks on the second layer and conduct data-driven selection of the 
parameters $\pi_j$ for different datasets. In detail,  $\pi_j$ is
selected through cross validation on a hold-out set in the learning process in
order to optimize the second-layer output of a specific task. Since
Target-Specific FECCM still has all the tasks instantiated on the second
layer, the re-training of the first-layer classifiers can still use data from
different datasets (i.e., with different task labels).
Here we train different models, i.e. different $\theta_{i}$'s and $\omega_{i}$'s, for different target tasks.  
\end{list}


\vspace*{\sectionReduceTop}
\section{\label{implementation}Scene Understanding: Implementation}
\vspace*{\sectionReduceBot}
\label{sec:implementation}

In this section we describe the implementation details of our
instantiation of \todo{FE-CCM} for scene understanding. 
Each of the classifiers described below for the sub-tasks are our
``base-model'' shown in Table~\ref{table:table_results}. In some sub-tasks,
our base-model will be simpler than the state-of-the-art models (that are
often hand-tuned for the specific sub-tasks respectively). However, even when
using base-models in our \todo{FE-CCM}, our model will still outperform the
state-of-the-art models for the respective sub-tasks (on the same standard
respective datasets) in Section~\ref{S.Results}. 

In order to explain the implementation details for the different tasks, we will
use the following notation. Let $i$ be the index of the tasks we consider.  We
consider 6 tasks for our experiments on scene understanding: scene
categorization ($i=1$), depth estimation ($i=2$), 
event categorization ($i=3$), saliency detection ($i=4$), object detection ($i=5$) and geometric 
labeling ($i=6$). The inputs for the $j^{th}$ task at the first layer are given by the low-level features $\Psi_{j}$. At the second layer, in addition to the original features $\Psi_{j}$, the inputs include the outputs from the first layer classifiers. This is given by,
\begin{equation}
\Phi_{j} = \left[ \Psi_{j} \  Z_1 \  Z_2 \  Z_3 \  Z_4 \  Z_5 \  Z_6  \right]
\label{eq:featurelayer2}
\end{equation}
where, $\Phi_{j}$ is the input feature vector for the $j^{th}$ task on the
second layer, and $Z_i$ ($i=1,\dots,6$) represents the output from the $i^{th}$
task which is appended to the input to the $j^{th}$ task on the second layer and so
on.

\smallskip
\noindent
\textbf{Scene Categorization.} 
For scene categorization, we classify an image into one of 
the 8 categories defined by
 Torralba et al.~\cite{Torralba_IJCV01}
  tall building, inside city, street, highway, coast, open country, mountain and forest.
We evaluate the performance by measuring the rate of incorrectly assigning a scene
label to an image on the MIT outdoor scene dataset \cite{Torralba_IJCV01}. 
The feature inputs for the first-layer scene classifier $\Psi_1 \in \Re^{512}$ is 
the GIST feature~\cite{Torralba_IJCV01}, extracted at $4\times4$ regions of the image, on 
4 scales and 8 orientations.

We use an RBF-Kernel SVM classifier \cite{torralba_website}, as the first-layer
scene classifier, and a multi-class logistic classifier for the second layer. 
The output of the first-layer scene classifier $Z_1 \in \Re^{8}$ is an
8-dimensional vector where each element represents the log-odds score of the
corresponding image belonging to a scene category. This 8-dimensional output is
fed to each of the second-layer classifiers.

\smallskip
\noindent
\textbf{Depth Estimation.} 
For the single image depth estimation task, we estimate the depth of every pixel in
an image. We evaluate the estimation performance by computing the root mean square error of the estimated depth with respect to ground truth laser scan depth using the Make3D Range Image dataset \cite{saxena_make3D,Saxena:NIPS2005}. 
We uniformly divide each image into $55\times305$ patches as ~\cite{saxena_make3D}. The feature inputs for the first-layer depth estimation $\Psi_2 \in \Re^{104}$ are features which capture
texture, color and gradient properties of the patch. This is obtained by convolving the image
with Laws' masks and computing the energy and Kurtosis over the
patch along with the shape features as described by Saxena et al.~\cite{saxena_make3D}. 

We use a linear regression for the first-level and second-level instantiation of the depth
estimation module. 
The output of the first-layer depth estimation $Z_2 \in \Re_{+}$ is the predicted depth of each patch in the image.
In order to feed the first-layer depth output to the second-layer classifiers, for the scene categorization and event categorization tasks, we use a vector with the predicted depth of all patches in the image; for the other tasks, we use the 1-dimensional predicted depth for the patch/pixel/bounding-box, etc.

\smallskip
\noindent
\textbf{Event Categorization:} 
For event categorization, we classify an image into one of the 8
sports events as defined by Li et al.~\cite{feifei2}: bocce, badminton, polo, 
rowing, snowboarding, croquet, sailing and rock-climbing.
For evaluation, we compute the rate 
of incorrectly assigning an event label to an image.
The feature inputs for the first-layer event classifier $\Psi_3 \in \Re^{43}$ is a 43-dimensional feature vector, which includes the top 30 PCA projections of the 512-dimensional
GIST features \cite{torralba_gist}, the 12-dimension global color features (mean and variance of RGB and YCrCb color channels
over the entire image), and a bias term.

We use a multi-class logistic classifier on each layer for event classification. 
The output of the first-layer event classifier $Z_3 \in \Re^{8}$ is an 8-dimensional vector where each element represents the log-odd score of the corresponding image belonging to a event category. This 8-dimensional output is feed to each of the second-layer classifiers.

\smallskip
\noindent
\textbf{Saliency Detection.}
The goal of the saliency detection task is to classify each pixel in the image as either salient or non-salient. 
We use the saliency detection dataset used by Achanta et. al. \cite{achanta_cvpr09} for our experiments. The feature inputs for the first-layer saliency classifier $\Psi_4 \in \Re^{4}$ includes the 3-dimensional color-offset features based on the Lab color space as described by Achanta et al.~\cite{achanta_cvpr09} and a bias term.

We use a logistic model for the saliency estimation classifiers on both layers. 
The output of the first-layer saliency classifier $Z_4 \in \Re$ is the log-odd score of a pixel being salient. 
In order to feed the first-layer saliency detection output to the second-layer classifiers, for the scene categorization and event categorization tasks, we form a vector with the predicted saliency of all the pixels in the image; for the other tasks, we use the 1-dimensional average saliency for the corresponding pixel/patch/bounding-box.

\smallskip
\noindent
\textbf{Object Detection.}
We consider the following object categories: car, person, horse and cow. 
We use the train-set and test-set of PASCAL 2006 \cite{pascal_voc_2006} for our
experiments. Our object detection module builds on the part-based detector of
Felzenszwalb et.~al.~\cite{voc_release4}. We first generate 5 to 100 candidate
windows for each image by applying the part-based detector with a low
threshold (over-detection). 
The feature inputs for the first-layer object detection classifier $\Psi_5 \in \Re^{K}$ are the HOG features extracted based on the candidate window as~\cite{dalal_cvpr05} plus the detection score from the part-based detector \cite{voc_release4}. $K$ depends on the number of scales to be considered and the size of the object template.

We learn an RBF-kernel SVM model as the first layer
classifier. The classifier assigns each window a $+1$ or $0$ label indicating
whether the window belongs to the object or not. For the second-layer
classifier, we learn a logistic model over the feature vector constituted by
the outputs of all first-level tasks and the original HOG feature. We use average precision
to quantitatively measure the performance. 
The output of the first-layer object detection classifier $Z_5\in \Re^{4}$ are the estimated 0 or 1 labels for a region to belong to the 4 object categories we consider. 
In order to feed the first-layer object detection output to the second-layer classifiers, we first generate a detection map for each object. Pixels inside the estimated positive boxes are labeled as ``+1'', otherwise they are labeled as ``0''. 
For scene categorization and event categorization on the second layer, we feed all the elements on the map; for the other tasks, we use the 1-dimensional average value on the map for the corresponding pixel/patch/bounding-box.

\smallskip
\noindent
\textbf{Geometric labeling.}
The geometric labeling task refers to assigning each pixel to one of three
geometric classes: support, vertical and sky, as defined by 
Hoiem et~al.~\cite{hoiem_IJCV}. 
For evaluation, we compute the accuracy of assigning the correct geometric label to a pixel.
The feature inputs for the first-layer geometry labeling classifier $\Psi_6 \in \Re^{52}$ are the region-based features as described by Hoiem et~al.~\cite{hoiem_IJCV}.

We use the dataset and the algorithm by
\cite{hoiem_IJCV} as the first-layer geometric labeling module. To
reduce the computation time, we avoid the multiple segmentations and instead
use a single segmentation with 100 segments per image. 
We use a logistic model as the second-layer classifier.
The output of the first-layer geometry classifier $Z_6 \in \Re^{3}$ is a 3-dimensional vector with each element representing the log-odd score of the corresponding pixel belonging to a geometric category. In order to feed the first-layer geometry output to the second-layer classifiers, for scene/event categorization we form a vector with the predicted scores of all pixels; for the other tasks we use the 3-dimensional vector with each element representing the average scores for the corresponding pixel/patch/bounding-box.

\vspace*{\sectionReduceTop}
\section{\label{S.Results}Experiments and Results}
\vspace*{\sectionReduceBot}

\addtocounter{footnote}{1}
\begin{table*}[t]
\vspace{-20pt}
\centering
\caption{\footnotesize{Summary of results for the SIX vision tasks. Our
method improves performance in every single task. (Note: Bold face corresponds to our model performing equally with or better than state-of-the-art.) }}
\vspace{-10pt}
\begin{scriptsize}
    \begin{tabular*}{1\linewidth}{@{}l | c c c c c |  c c c c | c }
    \hline
      & Event & Depth & Scene & Saliency & Geometric & \multicolumn{5}{c}{Object detection}\\
\cline{7-11}
    Model &  Categorization & Estimation  & Categorization & Detection & Labeling & Car$\;\;$ & Person$\;\;$ & Horse$\;\;$ & Cow $\;\;$ & Mean\\ 
     & (\% Accuracy)  & (RMSE in m)  &  (\% Accuracy)  &  (\% Accuracy)  &  (\% Accuracy)  &  \multicolumn{5}{c}{(\% Average precision)} \\ 
    \hline
    Images in testset  & 1579 & 400 & 2688 & 1000 & 300 & \multicolumn{5}{c}{2686}\\
\hline
Chance  & 22.5 & 24.6 & 22.5 & 50 & 33.3 & - & - & - & - &  -\\ 
\hline
\hline
    Our base-model  & 71.8 ($\pm$0.8) & 16.7 ($\pm$0.4) & 83.8 ($\pm$0.2)& 85.2 ($\pm$0.2) & 86.2 ($\pm$0.2)& 62.4 & 36.3 & 39.0 & 39.9 & 44.4\\
\hline
All-features-direct & 72.7 ($\pm$0.8)& 16.4 ($\pm$0.4)& 83.8 ($\pm$0.4) & 85.7 ($\pm$0.2)& 87.0 ($\pm$0.6) & 62.3 & 36.8 & 38.8 & 40.0 & 44.5\\
\hline
    State-of-the-art & 73.4 & 16.7 (MRF) \footnotemark[\value{footnote}]
    & 83.8 & 82.5 ($\pm$0.2) & 88.1 & 61.5 & 36.3 & 39.2 & 40.7 & 44.4\\
  model (reported) & Li~\cite{feifei2} & Saxena~\cite{saxena_make3D} & Torralba~\cite{torralba_website} & Achanta~\cite{achanta_cvpr09}
& Hoiem~\cite{hoiem_IJCV} & \multicolumn{5}{c}{Felzenswalb et.~al.~\cite{pedro} (base)}\\
\hline
    CCM ~\cite{ashutosh_nips} & 73.3 ($\pm$1.6) & 16.4 ($\pm$0.4) & 83.8 ($\pm$0.6) & 85.6 ($\pm$0.2) & 87.0 ($\pm$0.6)& 62.2 & 37.0 & 38.8 & 40.1 & 44.5\\
 (our implementation) & & & & & & & & &\\
\hline
\hline
    FE-CCM & & & & & & & & & &\\
(unified) & {\bf 74.3 ($\pm$0.6)} & {\bf 15.5 ($\pm$0.2)} & {\bf 85.9 ($\pm$0.3)} & {\bf 86.2 ($\pm$0.2)} & {\bf 88.6 ($\pm$0.2)} & {\bf 63.2} &  {\bf 37.6} & {\bf 40.1} & 40.5 & {\bf 45.4}\\
\hline
    FE-CCM & & & & & & & & & &\\
(one goal) & {\bf 74.2 ($\pm$0.8)} & {\bf 15.3 ($\pm$0.4)} & {\bf 85.8 ($\pm$0.5)} & {\bf 87.1 ($\pm$0.2)} & {\bf 88.6 ($\pm$0.3)} & {\bf 63.2} &  {\bf 37.9} & {\bf 40.1} & {\bf 40.7} & {\bf 45.5}\\
\hline
   FE-CCM & & & & & & & & & &\\
   (target specific) & 
{\bf 74.7 ($\pm$0.6)} & {\bf 15.2 ($\pm$0.2)} & {\bf 86.1 ($\pm$0.2)} & {\bf 87.6 ($\pm$0.2)} & { \bf 88.9 ($\pm$0.2)} & {\bf 63.2} &  {\bf 38.0} & {\bf 40.1} & {\bf 40.7} & {\bf 45.5}\\
 \bottomrule
   \end{tabular*}
\end{scriptsize}
 \label{table:table_results}
\end{table*}
\begin{figure*}[tb]
\vspace{-10pt}
\centering
         \includegraphics[height=5.5cm, width = 0.75 \linewidth]{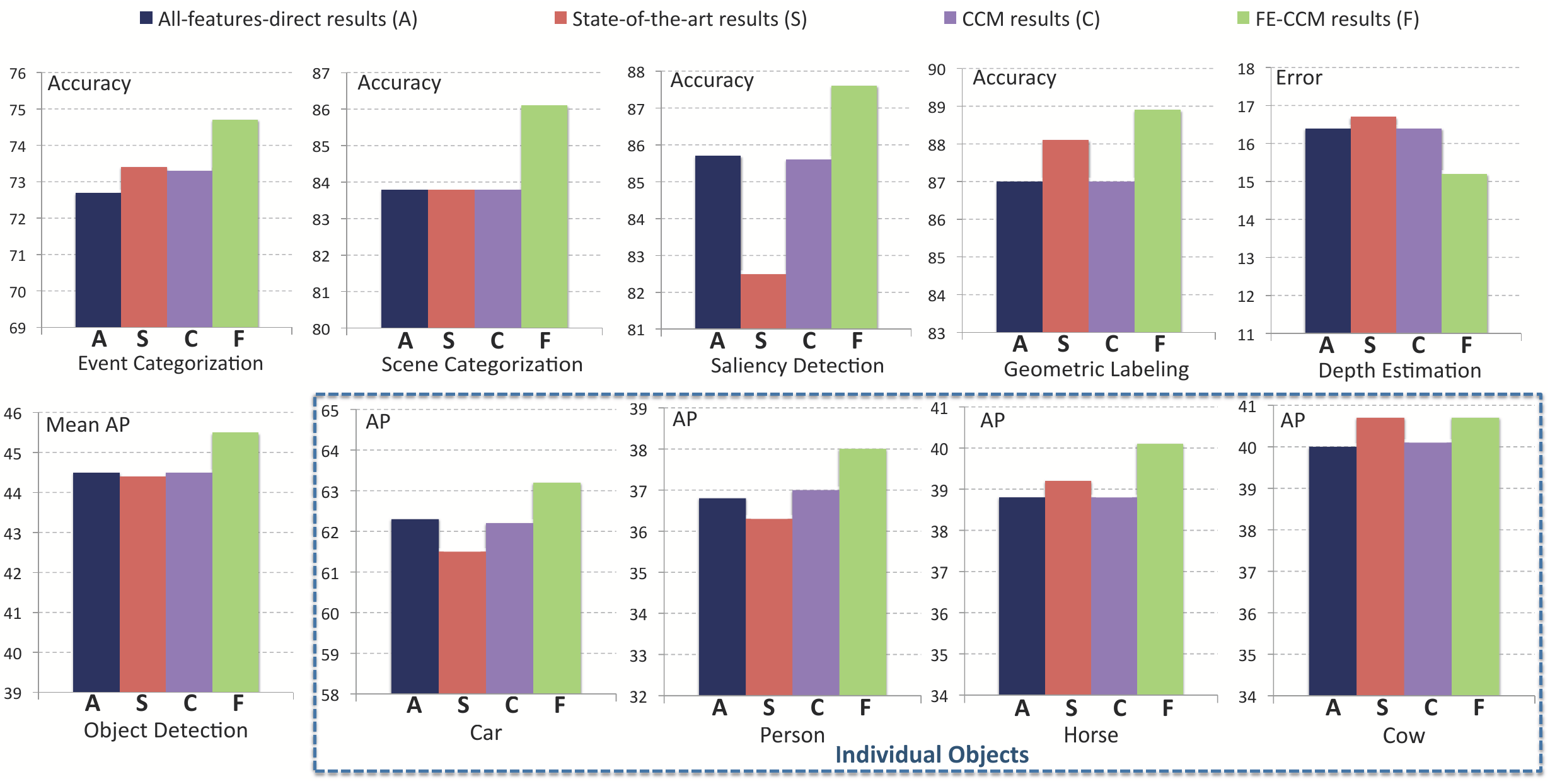}
         \label{fig:bar_graphs}\vspace{-10pt}         
\caption{\footnotesize{Results for the six tasks in scene understanding. Top: the performance for event categorization, scene categorization, saliency detection, geometric labeling, and depth estimation. Bottom: the average performance for object detection and the performance for the detection of individual object categories: car, person, horse, and cow. Each figure compares four methods: all-features-direct method, state-of-the-art methods, 
CCM, and  
the proposed FE-CCM method.}}
\label{fig:bar_graphs}
\vspace{-15pt}
\end{figure*}

\subsection{Experimental Setting}
\vspace*{\subsectionReduceBot}

The proposed FE-CCM model is a unified model which jointly optimizes 
for all the sub-tasks. We believe this is a powerful algorithm in that, while
independent efforts towards each sub-task have led to state-of-the-art
algorithms that require intricate modeling for that specific sub-task, the
proposed approach is a unified model which can beat the state-of-the-art
performance in each sub-task and, can be seamlessly applied across different applications.    

We evaluate our proposed method on combining six tasks introduced in Section~\ref{sec:implementation}. 
In our experiment, the training of FE-CCM takes 4-5 iterations.
For each of the sub-tasks in each of the domains, we evaluate our
performance on the standard dataset for that sub-task (and compare
against the specifically designed state-of-the-art algorithm for that dataset).
Note that, with such disjoint yet practical datasets, no image would have
ground truth available for more than one task. Our model handles this well.

In experiment we evaluate the following
algorithms as shown in Table~\ref{table:table_results},
\vspace{-7pt}
\begin{packed_item}
\item Base model: Our implementation (Section~\ref{implementation})
of each sub-task, which serves as a base model for our \todo{FE-CCM}.
(The base model uses less information than state-of-the-art algorithms for some
sub-tasks.) 
\item All-features-direct:  A classifier that takes all the
features of all
sub-tasks, appends them together, and builds a separate classifier for each sub-task.
\item State-of-the-art model: The state-of-the-art 
algorithm for each sub-task respectively on that specific dataset.
\item CCM: The cascaded classifier model by 
Heitz et~al.~\cite{ashutosh_nips}, which we re-implement for six sub-tasks.
\item FE-CCM (unified): This is our proposed model. Note that this is 
\textit{one single model}
which maximizes the joint likelihood of all the sub-tasks.
\item FE-CCM (one goal): In this case, we have only one sub-task instantiated on the second layer,
and the goal is to optimize the outputs of that sub-task. We train a specific one-goal FE-CCM for each sub-task. 
\item FE-CCM (target specific): In this case, we train a specific FE-CCM for each sub-task, by
using cross-validation to estimate $\pi_j$'s in
Equation~\ref{eq.forwardlayer1_heter}. Different values for $\pi_j$'s
result in different parameters learned for each FE-CCM.
\end{packed_item}
Note that both CCM and All-features-direct use information from all 
sub-tasks, and state-of-the-art models also use carefully designed
models that implicitly capture information from the other sub-tasks.

\vspace*{\subsectionReduceTop}
\subsection{Datasets}
\vspace*{\subsectionReduceBot}

The datasets used are mentioned in Section~\ref{implementation},
and the number of test images in each dataset is shown in
Table~\ref{table:table_results}.  For each dataset we use the same number of
training images as the state-of-the-art
algorithm (for comparison).  We perform 6-fold cross
validation on the whole model with 5 of 6 sub-tasks to evaluate the performance on
each task.  
We do not do cross-validation on object detection as it is standard on the PASCAL 2006
\cite{pascal_voc_2006} dataset (1277 train and 2686 test images respectively).

\vspace*{\subsectionReduceTop}
\subsection{Results}
\vspace*{\subsectionReduceBot}

\begin{figure*}[t]
\vspace{-10pt}
\centering
         {\includegraphics[height=9cm, width=0.8 \linewidth]{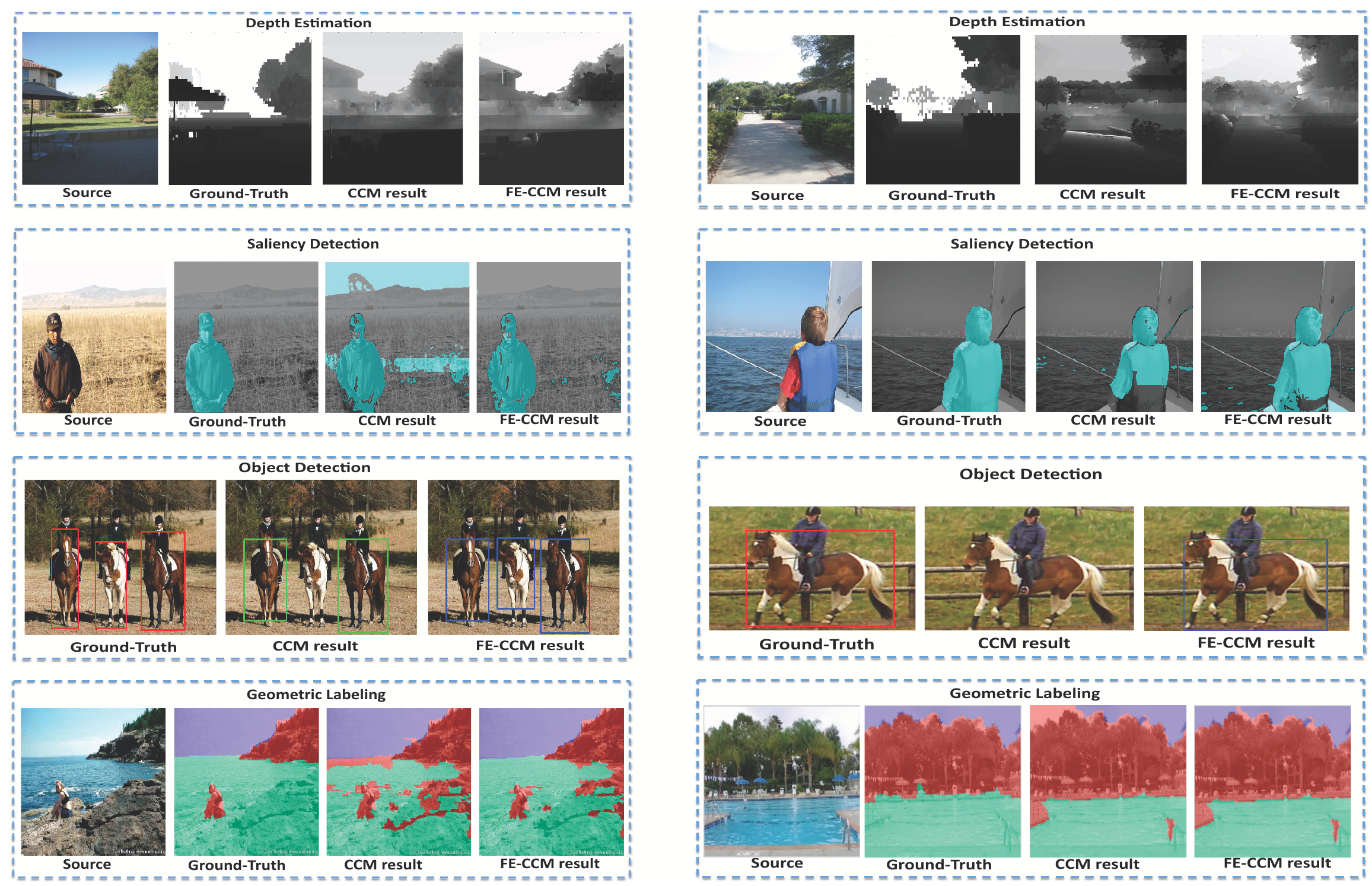}
         \label{fig:all_results}}
         \vspace{-10pt}
\caption{\footnotesize{Results showing improvement using the proposed model. From top to bottom: Depth estimation, Saliency detection, Object detection, Geometric labeling.
All depth maps in depth estimation are at the same scale (black means near and white means far); Salient region in saliency detection are indicated in cyan; Geometric labeling: Green=Support, Blue=Sky and Red=Vertical (Best viewed in color). 
}} 
\vspace{-12pt}
\label{fig:result_improvement}
\end{figure*}

To quantitatively evaluate our method for each of the sub-tasks, we consider
the metrics appropriate to each of the six tasks in
Section~\ref{implementation}.
Table~\ref{table:table_results} and Figure~\ref{fig:bar_graphs} show
that FE-CCM not only beats state of the art in \textit{all} the tasks but also does it
jointly as \textit{one single unified model}. 

In detail, 
we see that all-features-direct 
improves over the base model because it uses features from all the tasks.
The state-of-the-art classifiers improve on the base model by
explicitly hand-designing the task specific probabilistic model \cite{feifei2,saxena_make3D}
or by using adhoc methods to implicitly use information from other tasks \cite{hoiem_IJCV}.
Our FE-CCM model, which is a single model that was not given any manually designed
task-specific insight, achieves a more significant improvement over the
base model.

We also compare the three instantiations of FE-CCM in Table~\ref{table:table_results} (the last three rows). 
We observe that the target-specific FE-CCM achieves the best performance, by selecting a set of $\pi_{j}$'s to optimize for each task independently. Though the unified FE-CCM achieves slightly worse performance, it jointly optmizes for all the tasks by training only one set of parameters. The performance of one-goal FE-CCM is less stable compared to the other two instantiations. It is mainly because the first-layer classifiers only gain feedback from the specific task on the second layer in one-goal FE-CCM, which easily causes overfitting. 

\footnotetext[\value{footnote}]{The state-of-the-art method for depth estimation in~\cite{saxena_make3D} follows a slightly different testing procedure. In that case, our target-specific FE-CCM method achieves $RMSE=15.3$.}

We note that our target-specific FE-CCM, which is optimized for each task independently and
achieves the best performance, is a more fair comparison to the state-of-the-art because
each state-of-the-art model is trained specifically for the respective task.
Furthermore, Figure~\ref{fig:bar_graphs} shows the results for
CCM (which is a cascade without feedback information) and
all-features-direct (which uses features from all the tasks). This
indicates that the improvement is strictly due to the proposed feedback
and not just because of having more information.

\begin{figure}[t]
\centering
\includegraphics[height=2.7cm,width = 1.0 \linewidth]{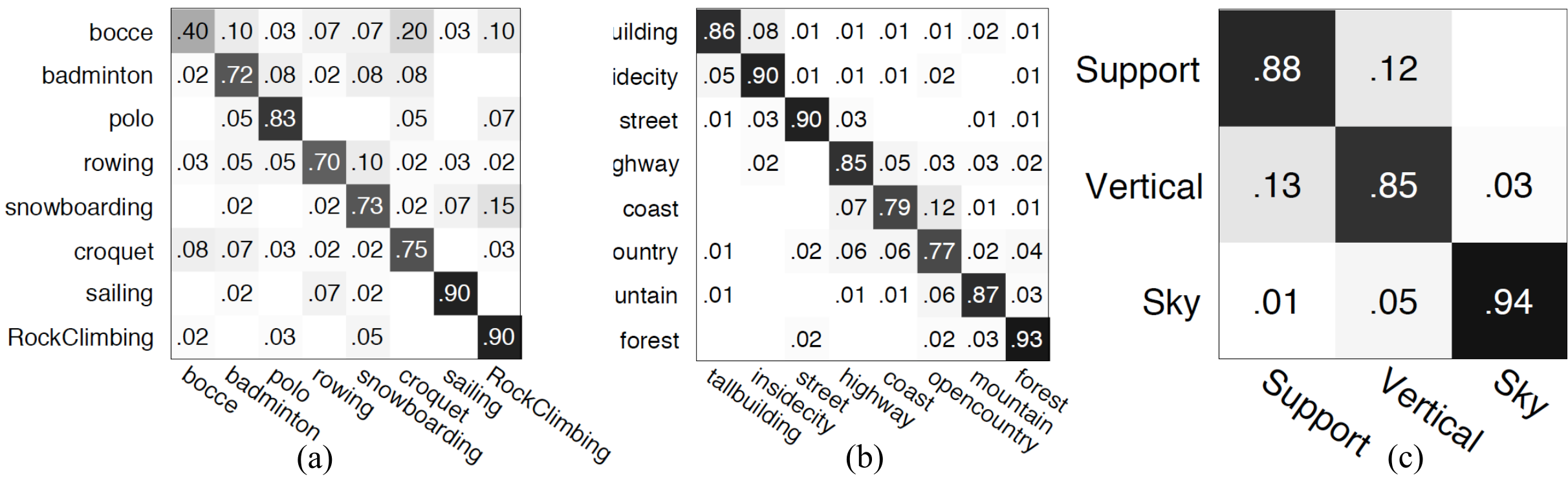}
\vspace{-20pt}
\caption{\footnotesize{Confusion matrix for (a) Event categorization; (b) Scene categorization; (c) Geometric labeling. All the results are gained with the proposed FE-CCM method. The average accuracy achieved by the proposed FE-CCM model outperforms the state-of-the-art methods for each of these tasks, as listed in Table~\ref{table:table_results}. }}
\label{fig:confusion_matrix}
\vspace{-15pt}
\end{figure}

\begin{figure*}[tb!]
\vspace{-15pt}
\centering
	\includegraphics[width = 0.8 \linewidth, height=1.4in]{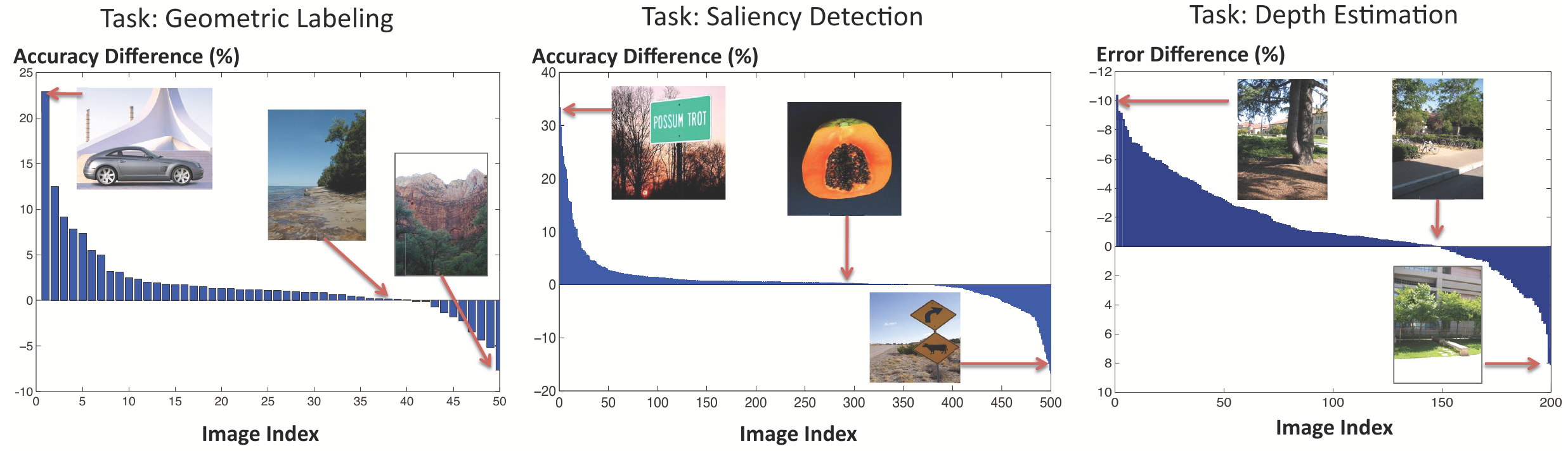}
\vspace{-10pt}
\caption{\footnotesize{Performance difference between the proposed unified FE-CCM method and the all-features-direct method for each test image, respectively for the tasks of geometric labeling, saliency detection, depth estimation, on one of the cross-validation folds. }}
\label{fig:results_perImage}
\vspace{-5pt}
\end{figure*}

We show some visual improvements due to the proposed FE-CCM in
Figure~\ref{fig:result_improvement}. In comparison to CCM, FE-CCM leads to better
depth estimation of the sky and the ground, and it leads to better coverage and
accurate labeling of the salient region in the image, and it also leads to 
better geometric labeling and object detection. Figure~\ref{fig:confusion_matrix}
also provides the confusion matrices for the three tasks: scene categorization, 
event categorization, geometric labeling.

\begin{figure*}[t]
\centering
        {\includegraphics[width=1.0 \linewidth, height=1.7in]{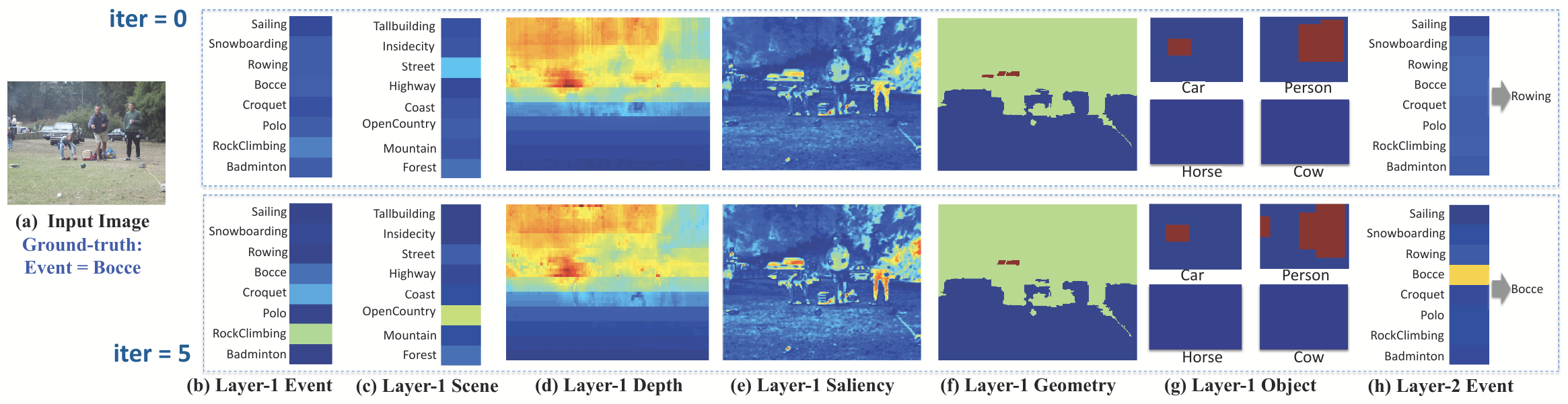}}
\vspace{-20pt}
\caption{\footnotesize{Illustration of the FE-CCM first-layer outputs for a single image.
(a) the input image from the sports-event dataset. Its groundtruth event label is ``Bocce''. (b-g)Outputs of the first-layer classifiers, at initialization (top row) and at the $5^{th}$ iteration (bottom row). (h) Outputs of the second-layer event classifier. Note that at initialization the first-layer classifiers are trained using ground-truth labels, i.e. the same as CCM.  In (b)(c)(d)(e)(h), Red=High-value, Blue=low-value. In (f), Blue=Ground, Green=Vertical, Red=Sky. In (g) Red=Object Presence. (Best viewed in color.)}}
\vspace{-18pt}    
\label{fig:firstlayer-output}
\end{figure*}

Figure~\ref{fig:results_perImage} provides scatter plots of the performance difference for each image between the unified FE-CCM method and the all-features-direct method, respectively for the tasks of geometric labeling, saliency detection, and depth estimation. We note that for all three tasks, the unified FE-CCM outperforms the all-features-direct method on most images. For geometric labeling and saliency detection, the improvement from the unified FE-CCM method is mainly due to large improvements on some images. For depth estimation, the improvement is scattered over many images.

\begin{table}[t]
\centering
\caption{\footnotesize{Summary of results for combining scene categorization and object detection, with partially-labeled datasets and fully-labeled datasets.}} 
\vskip -.15in
\begin{scriptsize}
    \begin{tabular}{@{}l |c c}
    \hline
      & Scene Categorization & Object Detection \\
    Model & (\% accuracy)  & (\% mean AP)  \\ 
     & partial-labeled / full-labeled & partial-labeled / full-labeled \\ 
\hline
\hline
Our base-model & 45.6 / 47.5 & 67.6 / 70.7\\
\hline
    All features direct & 46.8 / 49.1 & 71.2 / 72.5\\
\hline
    CCM \cite{ashutosh_nips} & 50.8 / 52.3 & 74.0 / 76.1\\
\hline
    {\bf FE-CCM (unified)} & {\bf 54.2 / 54.8} & {\bf 77.5 / 77.9}\\
 \hline
   \end{tabular}
\end{scriptsize}
 \label{T.labeleffects}
\vspace{-20pt}
\end{table}

\smallskip
\noindent
\textbf{The cause of improvement.} We have shown improvements of FE-CCM in Table~\ref{table:table_results} under the situation of heterogeneous datasets. The improvement can be caused by one or both of the following reasons: (1) the feedback process finds better error modes for the first-layer classifiers; (2) the feedback generates additional ``labels'' to retrain the first-layer classifiers. In order to analyze this, we consider the two tasks of scene recognition and object detection on the DS1 dataset in~\cite{ashutosh_nips}, which contains ground-truth labels for both the tasks. We compare the various methods under two settings: (1) train with the fully-labeled data; (2) train with  only the scene labels for one half of the training data and only the object labels for the second half. Table~\ref{T.labeleffects} compares the performance of training with partially-labeled datasets and the performance of different methods under these two settings. The experiments are performed using 5-fold cross validation. The unified FE-CCM method outperforms the other methods under both partially-labeled and fully-labeled situations. We note that all methods listed perform better when full labels are provided. In fact, FE-CCM achieves close performance in both settings. We also note that the FE-CCM method trained with partially-labeled datasets outperforms the CCM method trained with fully-labeled datasets, which indicates that the improvement achieved by the FE-CCM method is not simply from generating more labels for training the first-layer classifiers, but also due to finding useful modes for the first-layer classifiers. 

Figure \ref{fig:firstlayer-output} illustrates the first-layer outputs of a test image, respectively at initialization and at the $5^{th}$ iteration. Our initialization is the same as CCM, i.e., using ground-truth labels to train the first-layer classifiers. We note that with feedback, the first-layer output shifts to focus on more meaningful modes, e.g., At initialization, the event classifier has widespread confusion with other categories. With feedback, the event classifier turns to be confused with only the 'rock-climbing' and 'croquet' events which are more similar to 'bocce'. Moreover, the first-layer scene, depth, and object classifiers also give more meaningful predictions while trained with feedback. With better first-layer predictions, our FE-CCM correctly classifies the event as 'bocce', while CCM misclassifies it as 'rowing'.

\vspace*{\subsectionReduceTop}
\subsection{Discussion}
\vspace*{\subsectionReduceBot}
\begin{figure*}[t]
\vspace{-25pt}
\centering
        {\includegraphics[height=2.1in, width=0.95 \linewidth]{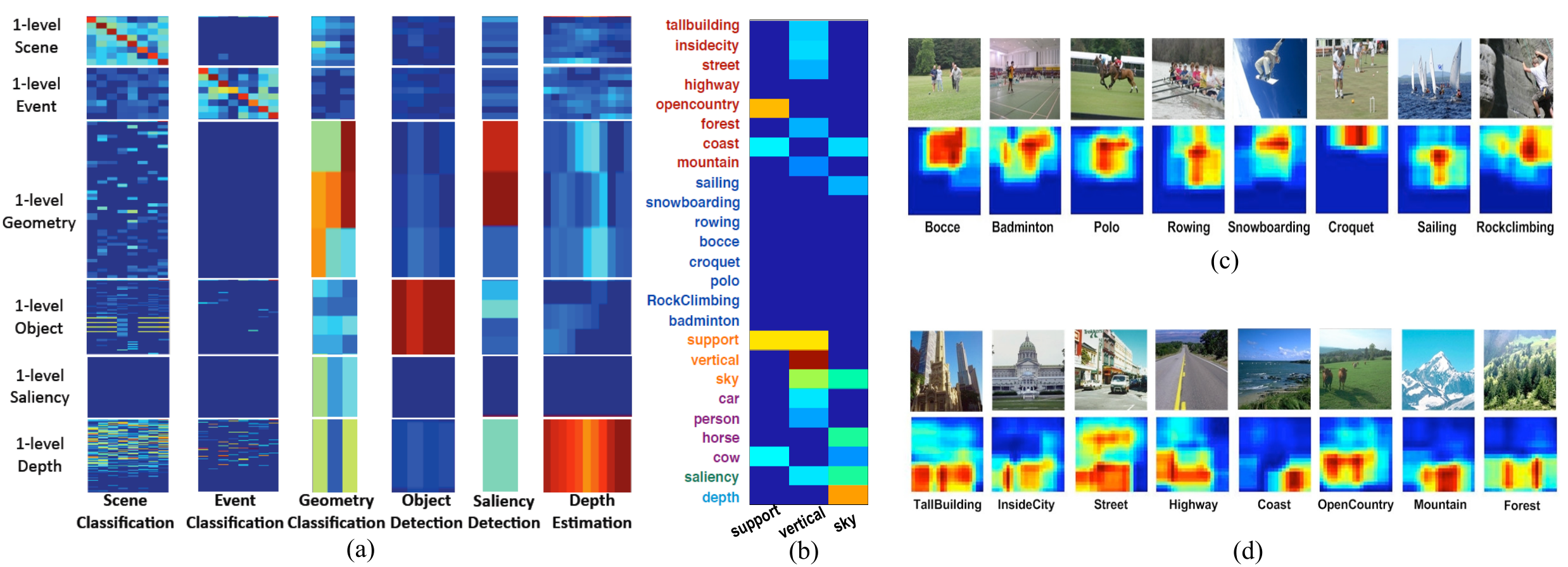}}
\vspace{-10pt}
\caption{\footnotesize{(a) The \emph{absolute} values of the weight
vectors for second-level classifiers, i.e. $\omega$. Each column shows the
contribution of the various tasks towards a certain task. (b) Detailed
illustration of the \emph{positive} values in the weight vector for a second-level
geometric classifier. (c)(d) Illustration of the importance of depths in different regions for predicting different
events (c) and scenes (d). An example image for each class is also shown above the map of the weights. (Note: Blue is low and Red is high. Best viewed in Color). }} 
\vspace{-6pt}    
\label{fig:weights}
\end{figure*}

FE-CCM allows each classifier in the second layer to learn which information
from the other first-layer sub-tasks is useful,
and this can be seen in the learned weights $\Omega$ for the second-layer. 
We provide a visualization of the weights for the six vision
tasks in Figure \ref{fig:weights}(a). We see that the model agrees with our
intuitions that large weights are assigned to the outputs of the same task from
the first layer classifier (see the large weights assigned to the diagonals in
the categorization tasks), though saliency detection is an exception which
depends more on its original features (not shown here) and the geometric
labeling output. We also observe that the weights are sparse. This is an advantage of our
approach since the algorithm automatically figures out which outputs from the first level classifiers
are useful for the second level classifier to achieve the best performance.

Figure \ref{fig:weights}(b) provides a closer look to the positive weights
given to the various outputs for a second-level geometric classifier. We
observe that large positive weights are assigned to
``mountain'', ``forest'', ``tall building'', etc.~for supporting the geometric
class ``vertical'', and similarly ``coast'', ``sailing'' and
``depth'' for supporting the ``sky'' class. These illustrate some of the
relationships the model learns automatically without any manual intricate
modeling.

Figure \ref{fig:weights}(c)
visualizes the weights given to the depth attributes (first-layer depth outputs)
for the task of event categorization.  Figure \ref{fig:weights}(d)
shows the same for the task of scene categorization. 
We see that the depth plays an important role in these tasks.
In Figure \ref{fig:weights}(c), we observe that most event categories 
rely on the middle part of the image,
where the main objects of the event are often located. E.g., most of the
``polo'' images have horses and people in the middle of the
image while many ``snowboarding'' images have people jumping in the
upper-middle part. 
For scene categorization, most of the scene categories (e.g., coast, mountain,
open country) have sky in the top part, which is not as discriminative
as the bottom part.
In scene categories of tall buildings and street, the upper part of the
street consists of buildings, which discriminates these two categories from the others.
Not surprisingly, our method had automatically figured this out (see
Figure~\ref{fig:weights}(d)).

\noindent
\textbf{Stability of the FE-CCM algorithm:} In this paper, we have presented
results for six sub-tasks. In order to find out how our method
scales with different combination and number of sub-tasks, we have tried 
several combinations, and in each case we get consistent improvement
in each sub-task. For example, in our preliminary experiments, we combined
depth estimation and scene categorization and our reduction in error are 12.0\%
and 13.2\% respectively.  
Combining scene categorization and object detection gives us 15.4\% and
10.2\% respective improvements (Table \ref{T.labeleffects}).
We then combined four tasks: event categorization,
scene categorization, depth estimation, and saliency detection, and got
improvements in all these sub-tasks \cite{feccm_eccv}.
Finally, we also combined different tasks for robotic applications,
and the performance improvement was similar.


\vspace*{\sectionReduceTop}
\section{Robotic Applications}
\label{Robotics}
\vspace*{\sectionReduceBot}
In order to show the applicability of our FE-CCM to different
scene understanding domains, we also used the proposed method in multiple
robotic applications. 

\vspace*{\subsectionReduceTop}
\subsection{Robotic Grasping}
\begin{figure}[t]
\vspace{-7pt}
\centering
    \includegraphics[width=0.8\columnwidth]{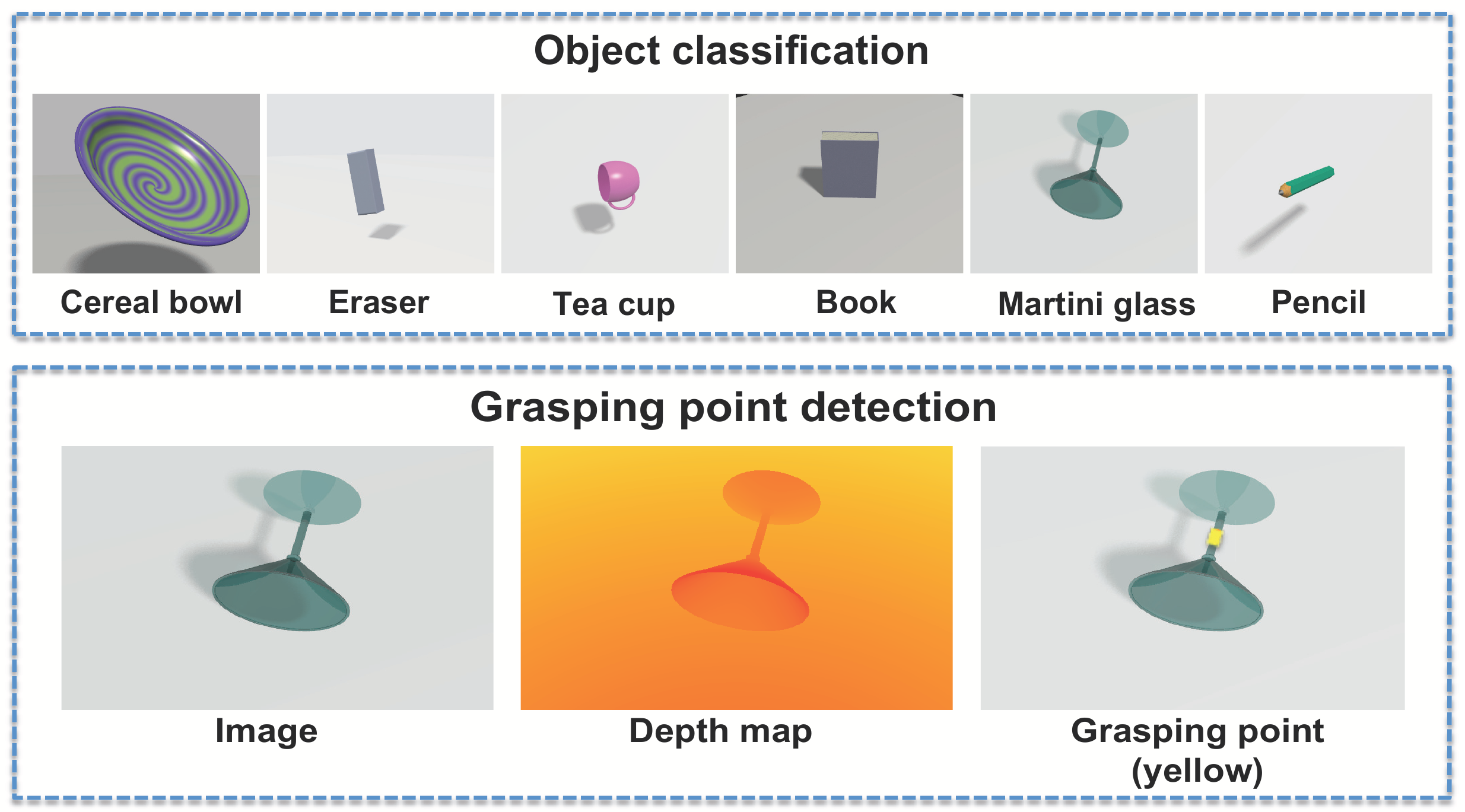}
\vspace{-10pt}
\caption{\footnotesize{Examples in the dataset used for the grasping robot experiments. The two tasks considered were a six-class, object classification task and grasping point detection task.}}
\vspace{-15pt}
\label{fig:grasping_dataset}
\end{figure}

\vspace*{\subsectionReduceBot}
Given an image and a depthmap (Figure~\ref{fig:grasping_dataset}), the goal of
the learning algorithm in a grasping robot is to select a point to
grasp the object (this location is called the grasp point,
\cite{Saxena:NIPSGrasping}).  It turns out that different categories of objects
demand different strategies for grasping. In prior work, Saxena et
al.~\cite{Saxena:NIPSGrasping,Saxena:IJRR} did not use object category information for
grasping.  In this work, we use our FE-CCM to combine object classification and
grasping point detection.

\noindent
\textbf{Implementation:}
We work with the labeled synthetic dataset by 
Saxena et~al.~\cite{Saxena:NIPSGrasping} which spans 6 object 
categories and also includes an aligned pixel level depth map for each image,
as shown in Figure \ref{fig:grasping_dataset}. 
The six object categories include spherically symmetric objects such as cerealbowl, 
rectangular objects such as eraser, martini glass, books, cups and long objects such as pencil. 

For grasp point detection, we compute image and depthmap features at 
each point in the image (using codes given by \cite{Saxena:NIPSGrasping}). 
The features describe the response of the image and 
the depth map to a bank of filters (similar to Make3D) while also
capturing information from the neighboring grid elements.
We then use a regression over the features. 
The output is a confidence score for
each point being a good grasping point. 
In an image, we pick the point with the highest score as the grasping point.

For object detection, we use a logistic classifier to perform the
classification. The output of the classifier is a 6-dimensional vector representing
the log-odds score for each category. The final classification is performed by
assigning the image to the category with the highest score.

\begin{table}[t]
\vspace{-10pt}
\centering
\caption{\footnotesize{Summary of results for the the robotic grasping experiment. Our
method improves performance in every single task.}} 
\vskip -.1in
\begin{scriptsize}
    \begin{tabular}{l | c c}
    \hline
      & Graping point & Object \\
    Model &  Detection & Classification \\ 
     & (\% accuracy)  & (\% accuracy)  \\ 
    \hline
    Images in testset  & 6000 & 1200 \\
\hline
Chance  & 50 & 16.7 \\ 
\hline
\hline
All features direct & 87.7 & 45.8\\
\hline
    Our base-model  & 87.7 & 45.8\\
\hline
    CCM (Heitz et.~al.) & 90.5 & 49.5\\
\hline
    {\bf FE-CCM} & {\bf 92.2} & {\bf 49.7}\\
 \hline
   \end{tabular}
\end{scriptsize}
 \label{T.grasping}
\vspace{-5pt}
\end{table}

\begin{figure}[t]
\vspace{-5pt}
\centering
    \includegraphics[height=0.7in]{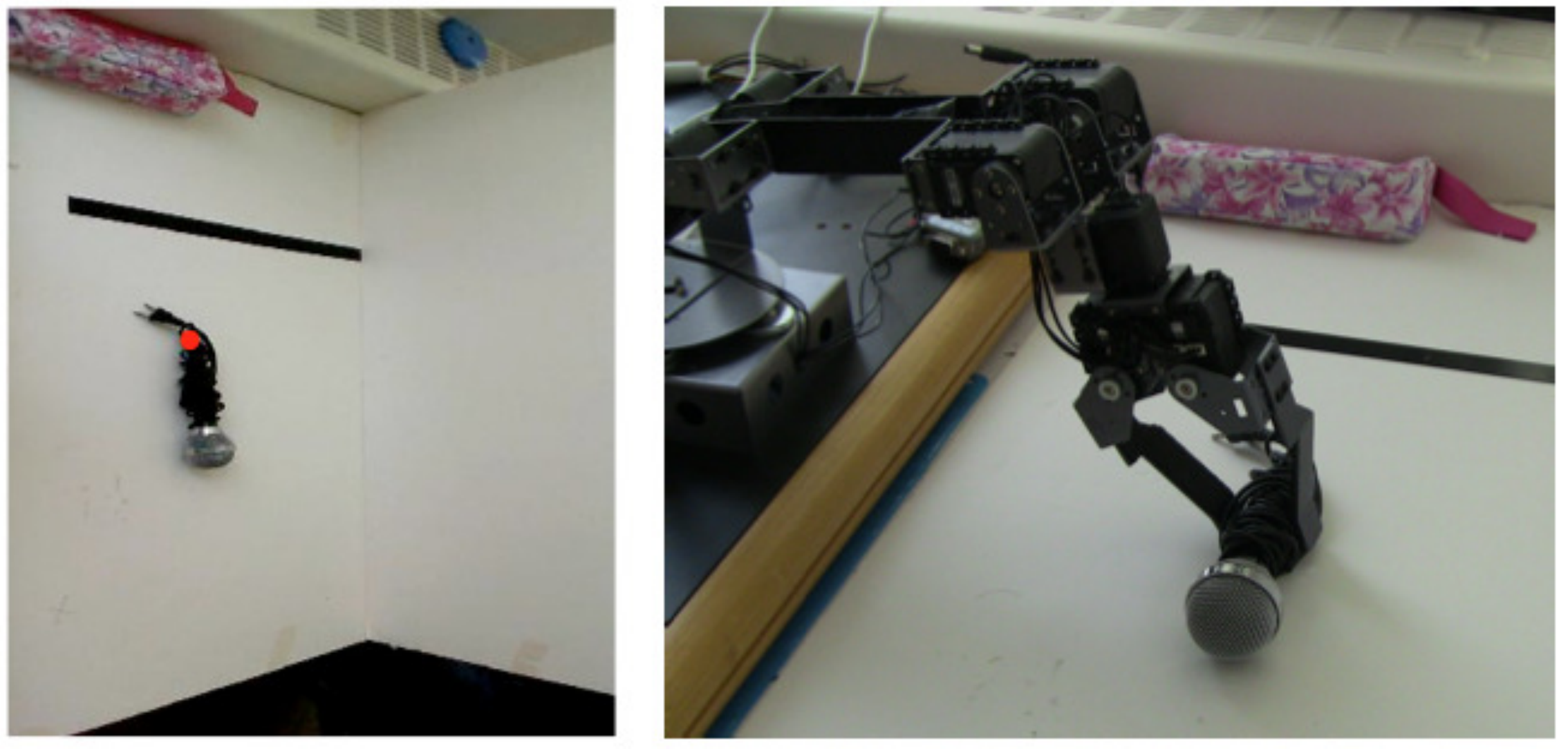}
\vspace{-10pt}
  \caption{\footnotesize{Left: the grasping point detected by our algorithm. Right: Our robot grasping an object using our algorithm.}}
\vspace{-20pt}
\label{fig.grasping}
\end{figure}

\noindent
\textbf{Results:}
We evaluate our algorithm on a dataset published in \cite{Saxena:NIPSGrasping},
and perform cross-validation to evaluate the performance on
each task. 
We use 6000 images for grasping point detection (3000 for training and 3000 for
testing) and 1200 images for object classification (600 for training and 600 for testing).
Table~\ref{T.grasping} shows the results for our algorithm's ability to 
predict the grasping point, given an image and the depths observed by the
robot using its sensors.  We see that our FE-CCM obtains significantly better
performance over all-features-direct and CCM (our implementation).  Figure~\ref{fig.grasping}
shows an example of our robot grasping an object.

\vspace*{\subsectionReduceTop}
\subsection{Object-finding Robot}
\vspace*{\subsectionReduceBot}

Given an image, the goal of an object-finding robot is 
to find a desired object in a cluttered room. As we have discussed 
earlier, some types of scenes such as living room are more likely to have
objects (e.g., shoes) than other types of scenes such as kitchen. Similarly, office
scenes are more likely to contain tv-monitors than kitchen scenes. 
Furthermore, it is also intuitive that shoes are more likely to appear on the
supportive surface such as floor, instead of the vertical surface such as the
wall.  Therefore, in this work, we use our FE-CCM to combine object detection
with indoor scene categorization and geometric labeling. 

\noindent
\textbf{Implementation:}
For scene categorization, we use the indoor scene subsets in the Cal-Scene
Dataset \cite{feifei_cvpr05} and classify an image into one of the four categories: bedroom,
living room, kitchen and office.  For geometric labeling, we use the Indoor Layout Data \cite{derek_iccv09} and assign each pixel to one of three geometry classes: ground, wall and ceiling. 
We use the same features and classifiers for scene categorization as in Section \ref{sec:implementation}.

For object detection, we use the PASCAL 2007 Dataset \cite{pascal2007} and our own shoe dataset to learn detectors 
for four object categories: shoe, dining table, tv-monitor, and sofa. We first use the part-based object detection algorithm in \cite{pedro} to create candidate windows, and then use the same classifiers as described in Section \ref{sec:implementation}. 

\begin{figure}[t]
\vspace{-15pt}
\centering
    \includegraphics[height = 0.8in]{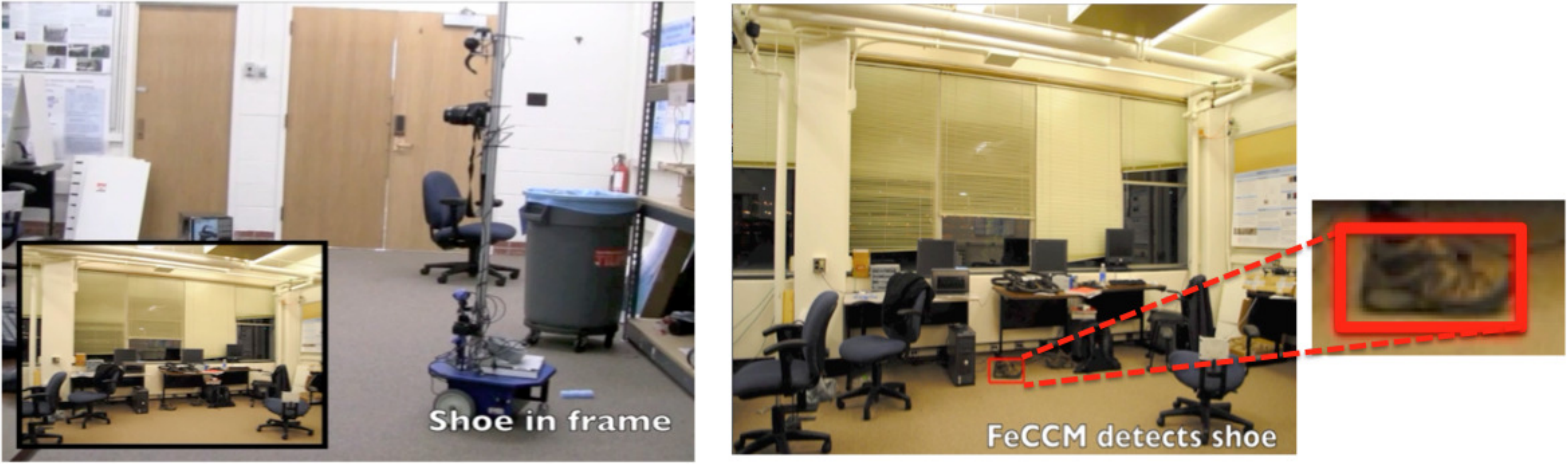}
\vspace{-5pt}
\caption{\footnotesize{Left: the shoe-finding robot, which has a camera to take photos of a scene. Right: the shoed detected using our algorithm.}}
\label{fig:shoefinding}
\vspace{-20pt}
\end{figure}

\noindent
\textbf{Results:}
We use this method to build a shoe-finding robot, as shown on 
Figure~\ref{fig:shoefinding}-left. With a limited number of training images (86
positive images in our case), it is hard to train a robust shoe detector to
find a shoe far away from the camera. However, using our FE-CCM model, the
robot learns to leverage the other tasks and performs more robust
shoe detection.  
Figure~\ref{fig:shoefinding}-right shows a successful detection.   For more details and videos, please
see~\cite{feccm_icra}.

\vspace*{\sectionReduceTop}
\section{\label{conclusions}Conclusions}
\vspace*{\sectionReduceBot}

We propose a method for combining existing classifiers for 
different but related tasks in scene understanding.  We only consider the individual
classifiers as a `black-box' (thus not needing to know the inner
workings of the classifier) and propose learning techniques for
combining them (thus not needing to know how to combine the tasks).  
Our method introduces feedback in the training process from
the later stage to the earlier one, so that a later classifier can provide
the earlier classifiers information about what error modes to focus
on, or what can be ignored without hurting the joint performance.

Our extensive experiments show that our unified model (a single FE-CCM trained for all the sub-tasks) 
improves performance significantly across \textit{all}
the sub-tasks considered over the respective state-of-the-art classifiers.
We show that this was the result of our feedback process. The classifier
actually learns meaningful relationships between the tasks automatically.
We believe that this is a small step towards holistic scene understanding.

\vspace{-10pt}
\ifCLASSOPTIONcompsoc
  \section*{Acknowledgments}
\else
  \section*{Acknowledgment}
\fi
\vspace{-5pt}
We thank Anish Nahar, Matthew Cong, TP Wong, Norris Xu, 
and Colin Ponce for help with the robotic experiments. We also thank 
Daphne Koller for useful discussions.

\ifCLASSOPTIONcaptionsoff
  \newpage
\fi

{\footnotesize
\vspace{-5pt}
\bibliographystyle{IEEEtran}
\bibliography{pami_feccm}
}

\vskip -80pt
\begin{biography}[\vskip -10pt{\includegraphics[width=0.7in,height=1.0in,clip]{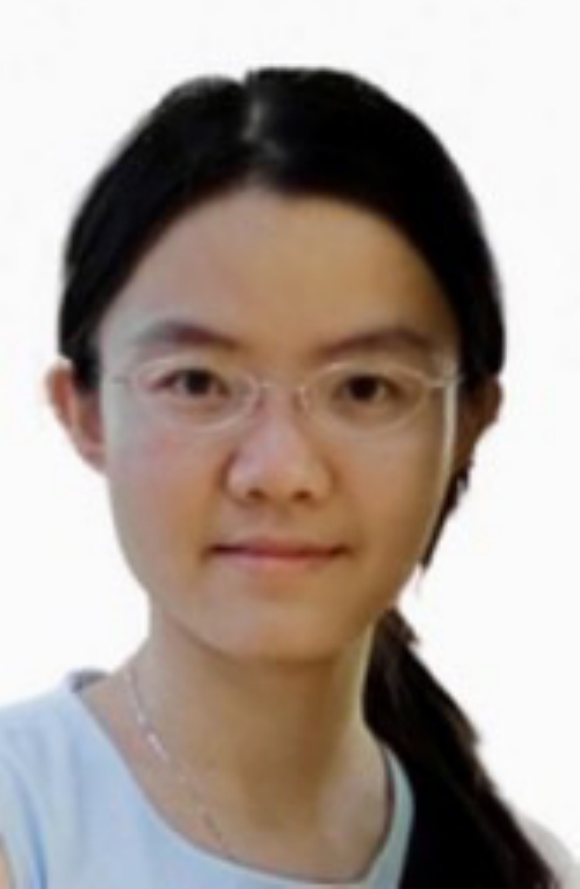}}]{Congcong Li}
is a Ph.D. candidate in the School of Electrical and Computer Engineering, Cornell University. She received her B.E. in 2005 and M.S. in 2007 from Tsinghua University, China. She started her Ph.D. study in Carnegie Mellon University, from 2007 and moved to Cornell University, in 2009. Her research interests include computer vision, machine learning, and image analysis.
\end{biography}
\vskip -80pt
\begin{biography}[\vskip -10pt{\includegraphics[width=0.7in,height=1.0in,clip]{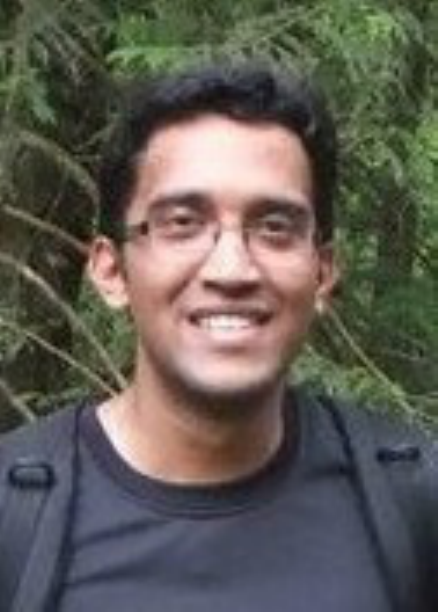}}]{Adarsh Kowdle}
is a Ph.D. student in the School of Electrical and Computer Engineering, Cornell University. He received his B.E. from Rashtreeya Vidyalaya College of Engineering (India) in 2007. He worked with Ittiam Systems (India) from June 2007 to July 2008. Before starting his Ph.D. in Cornell University in 2009, he spent a semester at Carnegie Mellon University.
His research interests include interactive computer vision algorithms and machine learning.
\end{biography}
\vskip -70pt
\begin{biography}[\vskip -10pt{\includegraphics[width=0.7in,height=1.0in,clip]{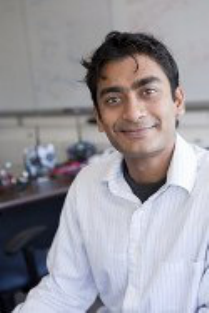}}]{Ashutosh Saxena}
is an assistant professor in computer science
department at Cornell University. His research interests include
machine learning, robotics and computer vision. He received his MS
in 2006 and Ph.D. in 2009 from Stanford University, and his B.Tech. in
2004 from Indian Institute of Technology (IIT) Kanpur.

Prof. Saxena has developed Make3D (http://make3d.cs.cornell.edu), an
algorithm that converts a single photograph into a 3D model. Tens of
thousands of users used this technology to convert their pictures to
3D. He has also developed algorithms that enable robots to perform
household chores such as load and unload items from a dishwasher. His
work has received substantial amount of attention in popular press,
including the front-page of New York Times, BBC, ABC, New Scientist
Discovery Science, and Wired Magazine. He has won best paper awards in
3DRR and IEEE ACE. He was also a recipient of National Talent Scholar
award in India, and Sloan research fellowship in 2011.
\end{biography}
\vskip -70pt
\begin{biography}[\vskip -10pt{\includegraphics[width=0.7in,height=1.0in,clip]{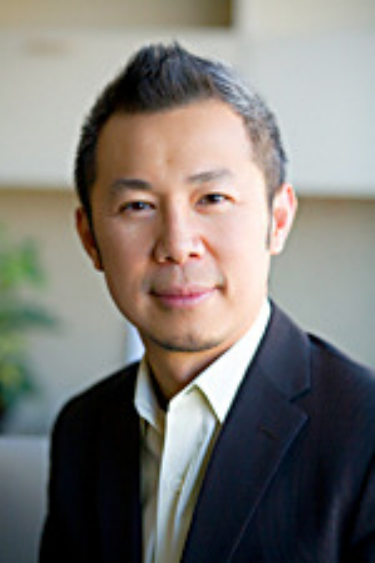}}]{Tsuhan Chen}
received the B.S. degree in electrical engineering from the National Taiwan 
University, Taipei, in 1987 and the M.S. and Ph.D. degrees in electrical engineering 
from the California Institute of Technology, Pasadena, in 1990 and 1993, respectively. 
He has been with the School of Electrical and Computer Engineering, 
Cornell University, Ithaca, NewYork, since January 2009, where he is Professor and Director. 
From October 1997 to December 2008, he was with the Department of Electrical and Computer Engineering, 
Carnegie Mellon University, Pittsburgh, PA, as Professor and Associate Department Head. 
From August 1993 to October 1997,he worked at AT\&T Bell Laboratories, Holmdel, NJ. He coedited a book 
titled Multimedia Systems, Standards, and Networks (Marcel Dekker, 2000). 

Prof. Chen served as the Editor-in-Chief for the IEEE Transactions on Multimedia in 2002 -- 2004. 
He also served in the Editorial Board of IEEE Signal Processing Magazine and 
as Associate Editor for IEEE Transactions on Circuits and Systems for Video Technology, 
IEEE Transactions on Image Processing, IEEE Transactions on Signal Processing, 
and IEEE Transactions on Multimedia. He received the Charles Wilts Prize at the 
California Institute of Technology in 1993. He was a recipient of the National 
Science Foundation Career Award, from 2000 to 2003. He received the Benjamin Richard Teare Teaching Award in 2006, 
and the Eta Kappa Nu Award for Outstanding Faculty Teaching in 2007. He was elected to the Board of Governors, 
IEEE Signal Processing Society, 2007 -- 2009, and a Distinguished Lecturer, 
IEEE Signal Processing Society, 2007 -- 2008. He is a member of the Phi Tau Phi Scholastic Honor Society.

\end{biography}
\vskip -70pt

\end{document}